\PassOptionsToPackage{table,dvipsnames}{xcolor}
\documentclass{maskwam}

\usepackage{amsmath,amssymb,amsfonts,mathtools}
\usepackage{algorithm,algorithmic}
\usepackage{wrapfig,tabularx,array,titletoc}
\usepackage{float,enumitem}
\newcommand{\rankbox}[2]{%
  \begingroup\setlength{\fboxsep}{0.55pt}\colorbox{#1}{\strut #2}\endgroup}
\newcommand{\first}[1]{\rankbox{orange!95}{\textbf{#1}}}
\newcommand{\second}[1]{\rankbox{orange!50}{#1}}
\newcommand{\third}[1]{\rankbox{orange!15}{#1}}
\newcommand{\firstcap}[1]{\rankbox{orange!95}{\textbf{#1}}}
\newcommand{\secondcap}[1]{\rankbox{orange!50}{#1}}
\newcommand{\thirdcap}[1]{\rankbox{orange!15}{#1}}
\newcolumntype{Y}{>{\centering\arraybackslash}X}
\newcommand{\sm}[2]{\mbox{#1}\hspace{1.8pt}\textcolor{black!55}{\textbar}\hspace{1.8pt}\mbox{#2}}
\newcommand{\metrichead}[1]{\shortstack{\textbf{#1}$\uparrow$\\[-0.8pt]{\fontsize{5.3pt}{5.3pt}\selectfont S\hspace{1.6pt}\textcolor{black!55}{\textbar}\hspace{1.6pt}M}}}
\newcommand{\humanref}[1]{\cellcolor{gray!18}#1}

\title{LeadVLA: Learning Follower-Aware Route Execution for Robot Leading}

\renewcommand{\authorlist}{{\sffamily\sbseries
 Tianle Zeng$^{1,2}$,\quad Hanjing Ye$^{2,\dagger,\ast}$,\quad Bingyi Xia$^{1}$,\quad Yuchen Zhou$^{2}$\\[0.3em]
 Zibo Zhang$^{3}$,\quad Yufei Huang$^{2,3}$,\quad Yonggen Ling$^{2,3}$,\quad Hong Zhang$^{1,\ast}$}}
\renewcommand{\affiliationlist}{{\normalsize\sffamily
 $^{1}$Southern University of Science and Technology\quad
 $^{2}$Tencent Robotics X\\[0.2em]
 $^{3}$Futian Laboratory}}
\renewcommand{\contributionlist}{{\small
 $^{\dagger}$Project leader.\quad
 $^{\ast}$Corresponding authors.}}

\abstract{\textbf{Abstract:}
Robot leading requires executing a route while continuously adapting to a designated follower.
Existing robot-leading systems often mediate follower understanding and physical route execution through explicit states, planners, or modular interfaces; directly learning this coupling through end-to-end trajectory prediction can instead entangle route execution with follower response.
We introduce LeadVLA, a structured multimodal policy that preserves the route as an explicit Path reference while learning a Pace for route progress and a Residual for local follower- and scene-conditioned adaptation.
To make this problem trainable and measurable, we develop LeadInfra, a configurable Unreal Engine-based closed-loop data-generation infrastructure from which we instantiate an approximately 4M-sample training corpus, and LeadBench, a standardized benchmark with 108 fully held-out closed-loop episodes.
LeadVLA achieves the highest task success (39.6\%) among evaluated robot policies on LeadBench while maintaining strong route fidelity and follower responsiveness, and transfers to unseen real-world environments without real-world policy fine-tuning.

}
\renewcommand{\metadatalist}{%
  \noindent
  {\small\sffamily\bfseries Project page: }
  \url{https://louiszengCN.github.io/LeadVLA_PAGE/}
  \par\vspace{2mm}
  \noindent
  {\small\sffamily\bfseries Correspondence: }
  \href{mailto:hankinye@tencent.com}{hankinye@tencent.com},
  \href{mailto:hzhang@sustech.edu.cn}{hzhang@sustech.edu.cn}
  \par
}

\begin{document}
\maketitle

\section{Introduction}
\label{sec:introduction}

Robot leading is an active human--robot navigation task in which a robot guides a designated follower along an intended route to a target destination, with applications in airports, retail spaces, campuses, and other public environments.
During leading, the follower may slow down, stop, fall behind, become temporarily unobservable, or briefly disengage, requiring the robot to adapt without losing the intended route.
The task therefore requires both reliable route execution and responsive human--robot interaction.

Existing robot-leading systems have made substantial progress on both aspects.
Classical methods typically rely on explicit geometric representations, interaction rules, and planning logic~\citep{fiore2015adaptive,xiao2021robotic,hwang2023system}, while learning-based approaches increasingly learn follower dynamics or guiding behavior from data~\citep{kim2023transforming,golemo2026milo}.
More recently, foundation models have introduced richer language grounding and semantic understanding~\citep{liu2024dragon,han2025space,hayamizu2026woofs}, yet follower understanding and physical route execution are still often connected through explicit states, planners, or modular interfaces~\citep{francis2025principles}.
In contrast, people naturally adjust their pace and local motion to a follower while preserving the intended route.
This motivates a direct embodied learning question: \textbf{can a robot learn a unified policy that executes a given route while adapting its actions to a designated follower?}

A natural intuition is to predict the complete leading trajectory end-to-end from route, visual, and language inputs.
However, we observe that this direct formulation can entangle route execution with follower response: changes in follower behavior that should mainly affect execution may also distort the underlying route geometry.
This motivates a simple structured formulation: \textbf{preserve the route as a stable execution reference while learning how to adapt its execution to the follower and local scene.}
Based on this idea, we introduce \textbf{LeadVLA}, which uses the given route as an explicit Path reference, a learned Pace to regulate progress along it, and a Residual for local follower- and scene-conditioned adaptation.

\begin{figure}[t!]
\centering
  \includegraphics[width=0.72\textwidth]{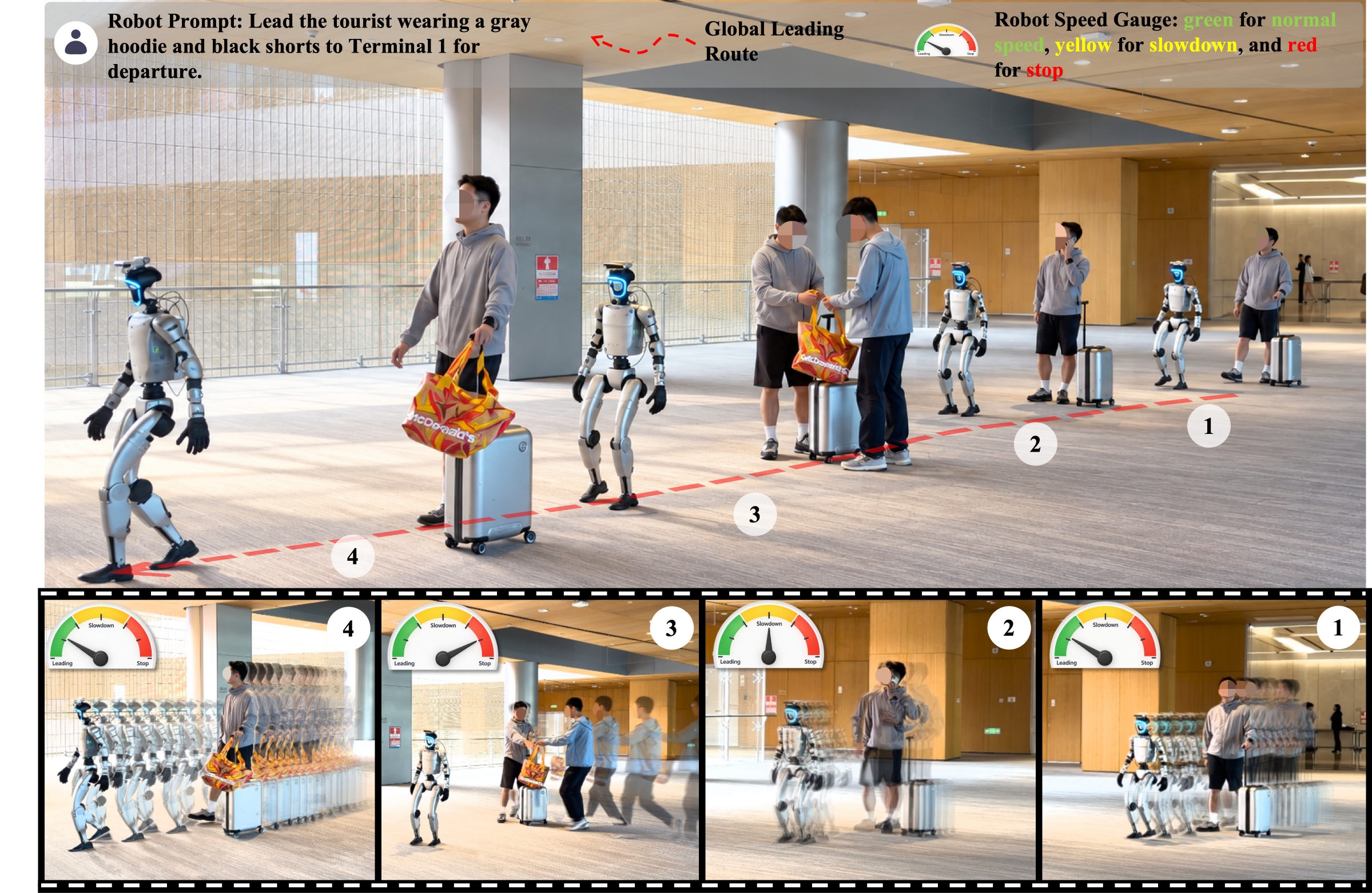}
  \caption{
    \textbf{Keep the route, adapt the execution.}
    A robot leads a language-designated tourist to the departure terminal along a planned route.
    The sequence shows (1) normal leading, (2) slowing down as the follower pauses for a phone call, (3) waiting while the follower briefly disengages to interact with another pedestrian, and (4) resuming afterward.
    The nominal route remains unchanged, while the robot adapts how it executes the route to the follower's evolving behavior.}
  \label{fig:teaser}
\end{figure}

Learning follower-aware route execution at scale also requires suitable training supervision and closed-loop evaluation.
Existing public resources do not yet jointly provide scalable action-level supervision and standardized evaluation tailored to this setting~\citep{wang2022tgrmpt,han2025space,kim2026guidedog}.
We therefore develop \textbf{LeadInfra}, a configurable Unreal Engine-based infrastructure that composes diverse routes, scenes, follower behaviors, and single- or multi-person configurations into closed-loop robot-leading episodes, together with \textbf{LeadBench}, a standardized closed-loop benchmark.
For this work, we instantiate an approximately 4M-sample training corpus with LeadInfra and evaluate on 108 fully held-out LeadBench episodes from disjoint environments, routes, actors, and scene instantiations under the same task definition and interaction contract.

On LeadBench, LeadVLA achieves the highest task success (39.6\%) among evaluated robot policies while maintaining strong route fidelity and follower responsiveness.
Real-world closed-loop experiments further show transfer to unseen environments without real-world policy fine-tuning.
Our main contributions are:

\textbf{(1)} We formulate robot leading as a follower-aware route-execution problem and introduce LeadVLA, which preserves the route as a stable execution reference while learning follower-aware adaptation.

\textbf{(2)} We develop LeadInfra and LeadBench, providing scalable closed-loop interaction supervision and standardized evaluation for follower-aware robot leading.

\textbf{(3)} We validate the formulation through controlled closed-loop experiments and real-robot deployment, and will release LeadVLA checkpoints, LeadInfra, and LeadBench to support continued research.

\section{Related Work}
\label{sec:related_work}

\textbf{Robot Leading Methods}.
Robot leading has progressed from explicitly designed human-aware navigation toward increasingly learned and foundation-model-based systems.
Early methods coupled route execution with follower state through geometric planners, interaction rules, or probabilistic models~\citep{fiore2015adaptive,xiao2021robotic,hwang2023system}.
Learning-based methods later learned follower dynamics or guiding policies from data, but typically relied on structured geometric or task representations~\citep{dewantara2016generation,kim2023transforming,kastner2022human,golemo2026milo}.
More recently, foundation models have introduced richer language grounding, spatial reasoning, and semantic understanding, while continuous physical execution often remains mediated by separate planning or control modules~\citep{liu2024dragon,song2025guide,han2025space,hayamizu2026woofs}.
LeadVLA instead studies how follower understanding and route execution can be coupled within a unified multimodal policy while retaining the route as an explicit execution reference.

\textbf{Embodied Navigation}.
Existing embodied navigation spans goal- or instruction-driven navigation, where the robot primarily acts toward a specified destination~\citep{zeng2025ezreal,batra2020objectnav,zeng2026taric}, and person following, where the observed person largely determines robot motion~\citep{ye2026refertrack, ye2025follow}.
Recent navigation foundation models further show that multimodal policies can directly predict navigation actions or trajectories from visual and language inputs~\citep{navfom,2026abotn1generalvisuallanguage}.
Robot leading differs in retaining its own route-level task intent while the designated follower primarily modulates \emph{how} that route is executed rather than defining the task-level destination or route.

\textbf{Data and Evaluation for Robot Leading}.
Existing resources cover human--guide interaction, person perception, and spatial or semantic understanding~\citep{hwang2026guidenav,wang2022tgrmpt,han2025space,kim2026guidedog,ye2025tpt}, while broader social-navigation simulators provide dynamic human environments and standardized evaluation~\citep{xia2026navisaaclab,puig2024habitat}.
Complete guide-robot systems have also been evaluated on route following, obstacle avoidance, and user-centered interaction~\citep{hwang2023system,kim2023transforming,golemo2026milo}.
However, these resources do not jointly provide scalable action-level supervision and a common closed-loop protocol for language-designated follower-aware route execution.
LeadInfra and LeadBench target this policy-learning and evaluation layer.

\section{Method}
\label{sec:method}

\subsection{Motivation and LeadVLA Overview}

\setlength{\intextsep}{2pt}
\setlength{\columnsep}{8pt}
\begin{wrapfigure}{R}{0.34\textwidth}
    \centering
    \includegraphics[width=\linewidth]{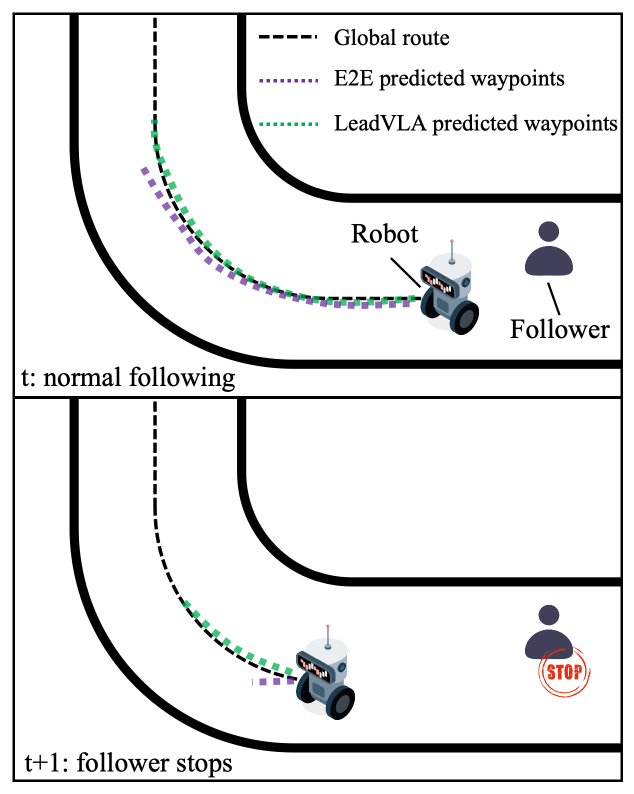}
    \caption{\textbf{Route--follower entanglement.}
    Direct trajectory prediction may distort the nominal route when reacting to follower behavior.}
    \label{fig:motivation}
\end{wrapfigure}

\textbf{Robot Leading Formulation.}
We consider robot leading, where a robot executes a given global route $P$ while adapting to a language-designated follower and the local scene.
At time $t$, the policy receives temporal front- and rear-view observations $\mathcal{O}_{1:t}$, a language instruction $\ell$ specifying who and how to lead, and a robot-local route context $\rho_t$ derived from $P$, and predicts an $H$-step robot-local trajectory $\hat{\tau}_{t,1:H}$, with $H=24$.

\textbf{Motivating Observation.}
Directly predicting the complete trajectory from these multimodal inputs can entangle route execution with follower response: changes that should mainly affect execution may also distort the underlying route geometry, particularly around turns (Fig.~\ref{fig:motivation}).
We therefore keep route-level task intent explicit while learning how its execution should adapt.

\textbf{LeadVLA Overview.}
LeadVLA represents the given route in two complementary robot-local forms.
The compact route context $\rho_t$ contains coordinate samples serialized into the VLM prompt, while a 24-step numerical Path reference $\mathcal{P}_t$ is retained outside the VLM for deterministic action composition.
The former supports reasoning about local route geometry, while the latter preserves the nominal route as an explicit execution reference.

Given $\mathcal{P}_t$, LeadVLA learns a scalar \emph{Pace} $\hat{a}_t$ and a horizon-wise \emph{Residual} $\hat{r}_{t,1:H}$.
Pace regulates progress along Path, enabling slowing, waiting, and resuming, while Residual provides local spatial correction for formation and scene-conditioned deviation.
Together, they adapt execution without discarding the route-level intent.

A single VLM forward pass produces $H$ shared horizon-wise action queries,
\begin{equation}
Q_t
=
\{q_{t,1},\ldots,q_{t,H}\}
=
f_{\theta}(\mathcal{O}_{1:t},\ell,\rho_t).
\label{eq:shared_action_queries}
\end{equation}
Pace and Residual are decoded from the shared action representation, with
$\bar{q}_t=\frac{1}{H}\sum_{k=1}^{H}q_{t,k}$,
$\hat{a}_t=g_{\mathrm{pace}}(\bar{q}_t)$, and
$\hat{r}_{t,1:H}=g_{\mathrm{res}}(Q_t)$,
where $g_{\mathrm{pace}}$ is a scalar Pace head and
$g_{\mathrm{res}}$ is a shared horizon-wise Residual decoder applied to the action queries.
The same representation is supervised with follower-related semantics, with selected spatial semantics additionally coupled to Residual (Sec.~\ref{sec:follower_representation}).

The deterministic Path Composer combines the explicit Path reference with the learned components:
\begin{equation}
\hat{\tau}_{t,1:H}
=
\mathcal{C}
\left(
\mathcal{P}_t,
\hat{a}_t,
\hat{r}_{t,1:H}
\right).
\label{eq:path_composer_overview}
\end{equation}

\begin{figure*}[t]
    \centering
    \includegraphics[width=\textwidth]{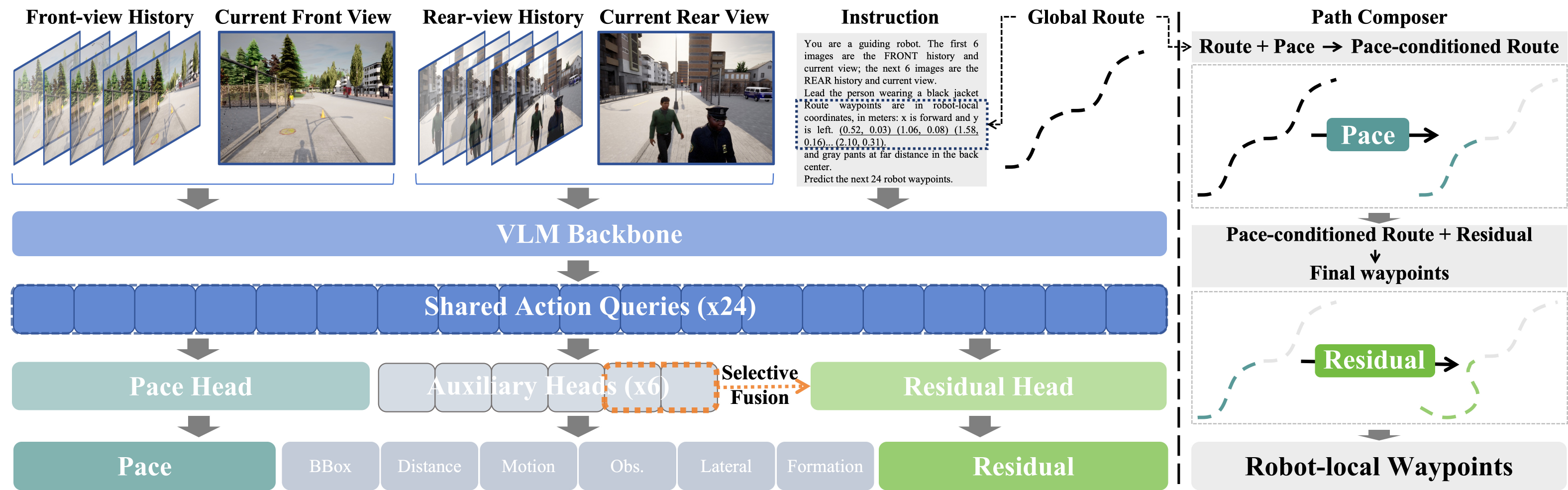}
    \caption{\textbf{Overview of LeadVLA.}
    Temporal front/rear observations, language, and route context produce 24 shared action queries.
    The route is also retained as an explicit Path reference.
    Pace regulates route progress, Residual provides local adaptation, and selected follower semantics couple only to Residual.
    A deterministic Path Composer outputs the final robot-local waypoints.}
    \label{fig:leadvla_overview}
\vspace{-15pt}
\end{figure*}

\subsection{Path Composer}

The Path Composer implements structured route execution by modulating route progress with Pace and adding Residual as a local correction.

\textbf{Robot-local Path Representation.}
Let $P(s)$ denote the global route pose at arc-length position $s$.
We parameterize the route by arc length so that execution progress is represented by a single scalar: changing $s$ advances along the same route geometry without directly modifying its spatial shape, including around turns.
At time $t$, the forward route segment is expressed in the current robot frame as
\begin{equation}
    \mathbf{p}_t(s)
    =
    \operatorname{Local}_t
    \left(
        P(s_t^{\mathrm{cur}}+s)
    \right),
\label{eq:local_path}
\end{equation}
where $s_t^{\mathrm{cur}}$ is the current route progress, $s$ is relative arc-length progress, and
$\operatorname{Local}_t(\cdot)$ transforms a global route pose into the robot-local frame at time $t$.
The Path reference is sampled as
$\mathcal{P}_t=\{\mathbf{p}_t(s_{t,k})\}_{k=1}^{H}$, with $s_{t,0}=0$.

\textbf{Pace-conditioned Path Composition.}
The predicted Pace $\hat{a}_t\in[0,1]$ specifies the target route-progress scale.
Starting from the previously executed Pace $a^{\mathrm{exec}}_{t-1}$, the Composer rate-limits the transition toward $\hat{a}_t$ to obtain a smooth horizon-wise profile $\hat{a}_{t,1:H}$.

This profile modulates the nominal arc-length increments:
\begin{equation}
    \Delta s_{t,j}
    =
    s_{t,j}-s_{t,j-1},
    \qquad
    \hat{s}_{t,k}
    =
    \sum_{j=1}^{k}
    \Delta s_{t,j}\hat{a}_{t,j}.
\label{eq:pace_progress}
\end{equation}
The corresponding waypoint is
$\tilde{\mathbf{p}}_{t,k}=\mathbf{p}_t(\hat{s}_{t,k})$.
Because Pace changes the arc-length positions queried on the Path rather than scaling robot-local coordinates, it changes how far the robot advances without deforming route geometry.
Small Pace values therefore slow or stop progress, while larger values smoothly resume leading.

\textbf{Residual Composition.}
Each horizon-wise Residual
$\hat{r}_{t,k}=(\Delta\hat{x}_{t,k},\Delta\hat{y}_{t,k},\Delta\hat{\psi}_{t,k})$
captures local motion not explained by route progress alone.
The final waypoint is
$\hat{\tau}_{t,k}=\tilde{\mathbf{p}}_{t,k}\oplus\hat{r}_{t,k}$,
where $\oplus$ denotes addition in robot-local position and wrapped heading composition.
Thus, Pace determines progress along the fixed route, while Residual provides local adaptation around the route-derived reference.
Implementation details are provided in Appendix~\ref{app:path_composer}.

\begin{figure}[t]
    \centering
    \includegraphics[width=\linewidth]{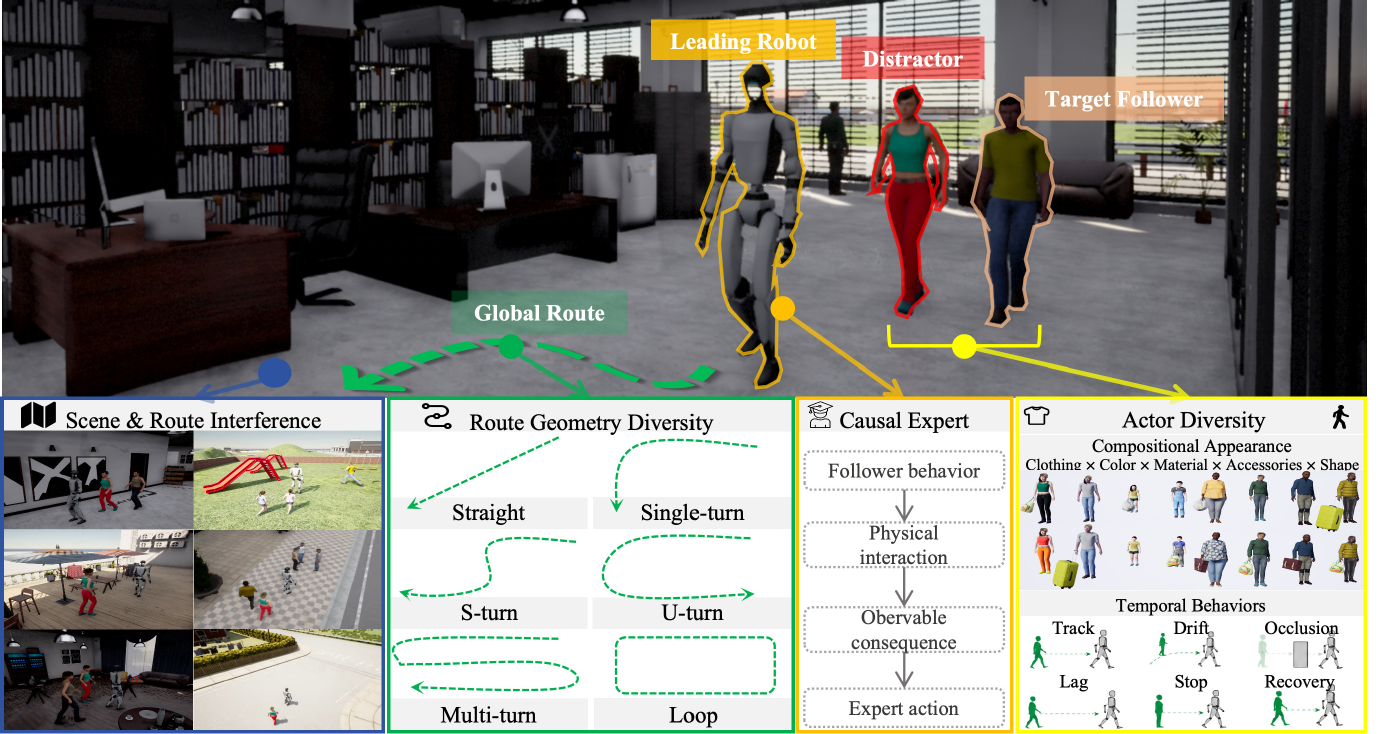}
    \caption{
    \textbf{Overview of LeadInfra.}
    LeadInfra compositionally combines route geometry, scenes, follower behaviors, appearances, and target--distractor configurations to generate diverse closed-loop robot-leading episodes.
    A causal expert converts realized interaction consequences into robot-action supervision without directly observing hidden behavior programs.
    }
    \label{fig:leadinfra}
    \vspace{-15pt}
\end{figure}

\subsection{Follower-Aware Representation and Selective Coupling}
\label{sec:follower_representation}

The structured action space defines \emph{how} adaptation is expressed; we next supervise \emph{what follower information} the shared action representation should capture.

We use six lightweight auxiliary heads.
Target localization supports grounding of the language-designated follower; lateral and distance capture relative geometry; observation and motion capture temporal presence and movement; and formation indicates whether the requested robot--follower relation is being acquired, maintained, or corrected.
These objectives supervise the shared representation $Q_t$ directly rather than forming an explicit follower-state vector for a separate controller.

\textbf{Selective Semantic Coupling.}
All six objectives shape $Q_t$, but only the predicted lateral and formation distributions are explicitly fused into Residual, as these cues are most directly aligned with local spatial correction.
Let $p_t^{\mathrm{lat}}$ and $p_t^{\mathrm{form}}$ denote the corresponding predicted distributions.
Selective coupling is implemented as
\begin{equation}
\Delta q_t =
F_{\mathrm{sem}}\!\left(p_t^{\mathrm{lat}},p_t^{\mathrm{form}}\right),
\quad
q^{\mathrm{res}}_{t,k}=q_{t,k}+\Delta q_t,
\quad
\hat{r}_{t,k}=g_{\mathrm{res}}\!\left(q^{\mathrm{res}}_{t,k}\right).
\label{eq:semantic_coupling}
\end{equation}

Pace remains decoded from the original pooled representation $\bar{q}_t$.
Target localization, distance, observation, and motion affect action only through supervision of the shared representation, rather than through direct semantic-to-action fusion.
This separates broad follower-aware representation learning from explicit coupling reserved for local spatial adaptation.

Detailed semantic definitions and coupling ablations are provided in Appendices~\ref{app:semantic_supervision} and~\ref{app:semantic_coupling}.

\subsection{Referent-Aware Co-training}
\label{sec:referent_cotraining}

In multi-person scenes, follower-aware control must also determine \emph{whose} behavior should influence action.
Because the follower is specified by language, the policy should respond to the designated person rather than nearby distractors.

\textbf{Counterfactual Referent Pairs.}
We preserve the route, scene, actors, and behavior while changing only which person is designated by language.
The same physical event can therefore require different actions: a designated follower stopping should induce slowing or waiting, whereas the same behavior from a distractor should not.
These matched pairs supervise target-conditioned action attribution.

\textbf{Dynamic Target Reassignment.}
In selected episodes, the language instruction changes the designated follower while both people remain in the scene.
The policy must then update which person's behavior controls subsequent execution rather than remain attached to the previous referent.
Together, counterfactual pairs supervise \emph{which person} should influence action, while reassignment requires this attribution to update online.

\textbf{Joint Training Objective.}
These interventions modify training supervision without introducing a referent-specific loss.
LeadVLA uses the same joint objective
$\mathcal{L}=\mathcal{L}_{\mathrm{act}}+\lambda_{\mathrm{sem}}\mathcal{L}_{\mathrm{sem}}$,
with $\lambda_{\mathrm{sem}}=0.25$.
The action objective supervises waypoint prediction, Pace, Residual, and regularization, while the semantic objective supervises the six auxiliary heads.
Detailed losses and training configuration are provided in Appendix~\ref{app:training}.

\subsection{LeadInfra and LeadBench}
\label{sec:leadinfra}

\textbf{LeadInfra.}
To provide scalable supervision for follower-aware route execution, we develop \textbf{LeadInfra}, a configurable infrastructure for generating closed-loop robot-leading data.
As illustrated in Fig.~\ref{fig:leadinfra}, LeadInfra composes route geometry, scene conditions, local interference, follower motion and visibility, requested formation, appearance, and single- or multi-person configurations.
Recombining these factors across routes and scenes encourages follower-conditioned execution rather than memorization of fixed trajectory patterns.

A key design is causal expert supervision from realized interaction evidence.
Follower behaviors are scripted, but their hidden labels are never exposed to the robot expert: for example, a scripted stop triggers no immediate response, and the expert reacts only after observable consequences such as increasing separation or sustained target loss emerge.
Robot actions are therefore supervised from realized interaction rather than hidden future behavior programs.

For this work, we instantiate approximately 4M training samples over roughly 1K nominal routes and more than 65K m$^2$ of navigable indoor and outdoor environments.
The corpus covers 8 follower-motion programs, 5 presence and reacquisition conditions, 6 language-conditioned formations, and single- and multi-person interactions, yielding 240 nominal motion--presence--formation combinations before additional route, appearance, referent, and scene variation.
Further details are provided in Appendix~\ref{app:leadinfra}.

\begin{figure*}[t]
    \centering
    \includegraphics[width=\textwidth]{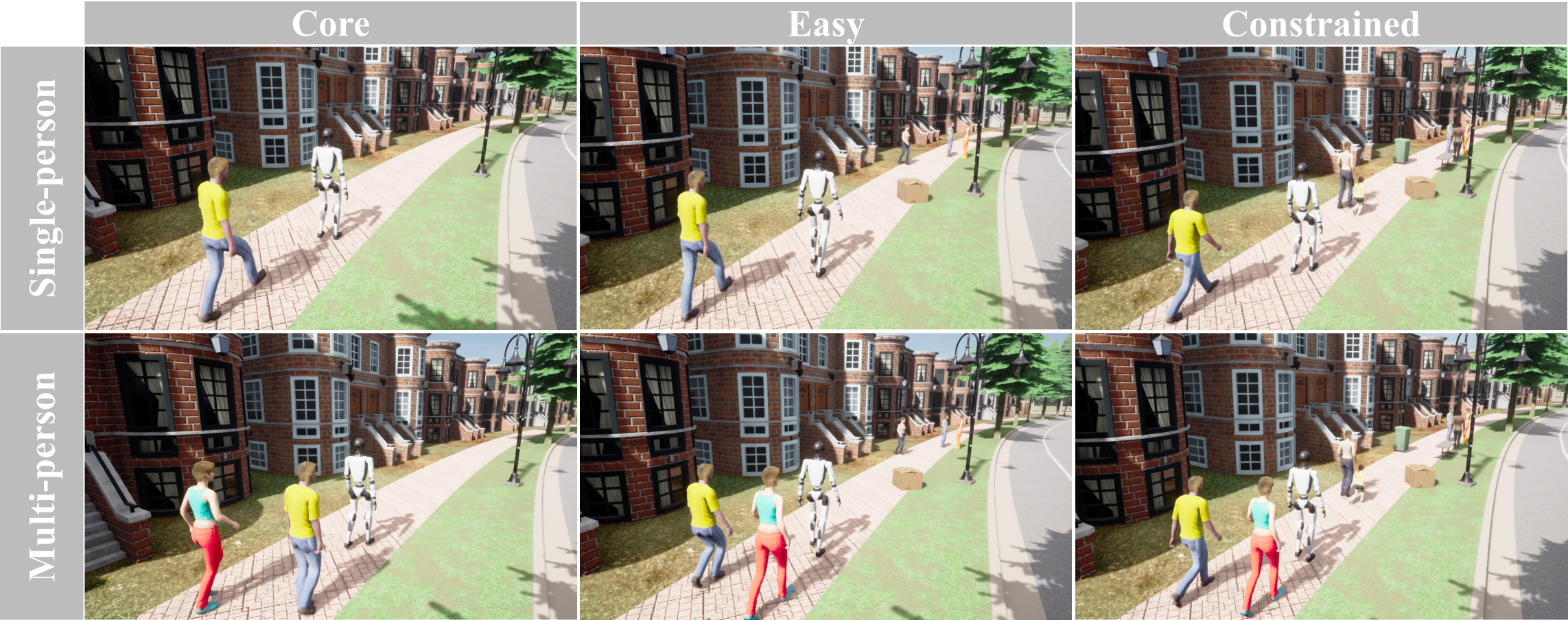}
    \caption{
\textbf{Progressive LeadBench evaluation.} Columns increase route-execution difficulty from clean conditions (Core) to moderate (Easy) and stronger interference (Constrained); rows compare single-person leading with multi-person cases containing distractors. The progression increasingly couples local route recovery with follower-aware control.
    }
    \label{fig:leadbench_difficulty}
 \vspace{-20pt}
\end{figure*}

\textbf{LeadBench.}
\label{sec:leadbench}
To provide standardized evaluation, we develop \textbf{LeadBench}, a frozen closed-loop benchmark in which all policies face the same routes, actors, interaction conditions, instructions, and evaluation events under a common policy-visible information boundary.
Each policy action changes subsequent states and observations; in multi-person cases, matched counterfactual rollouts keep the scene and human behavior fixed while changing only the language-designated follower.
LeadBench is disjoint from LeadInfra training in environments, routes, human actors, and scene instantiations while retaining the same task definition and interaction contract.

\textbf{Progressive Evaluation Conditions.}
LeadBench contains 108 held-out episodes across \textbf{Core}, \textbf{Easy}, and \textbf{Constrained}, which respectively introduce no additional, moderate, and stronger repeated route interference; each includes single- and multi-person settings.

\textbf{Evaluation Metrics.}
We report \textbf{Lead Success Rate (LSR)} for joint robot--follower completion, \textbf{Route Fidelity (RF)} for route preservation and forward progress, \textbf{Target Response Score (TRS)} for designated-follower response, and \textbf{Social Contract Score (SCS)} for instructed distance and formation.
Detailed definitions, aggregation rules, and termination conditions are provided in Appendix~\ref{app:leadbench}.

\section{Experiments}
\label{sec:experiments}

\subsection{Experimental Setup}
\label{sec:exp_setup}

\textbf{Evaluation protocol.}
We evaluate all methods in closed loop under the frozen LeadBench protocol.
All methods face the same predefined episode conditions, policy-visible information boundary, and common execution environment.
We report LSR, RF, TRS, and SCS as defined in Sec.~\ref{sec:leadbench}, with detailed metric definitions, aggregation rules, and termination conditions provided in Appendix~\ref{app:leadbench}.

\textbf{Baselines.}
We compare against representative robot-guiding paradigms.
A Fiore-style controller~\citep{fiore2015adaptive} represents classical modular guiding with explicit follower-state reasoning and geometric route execution.
\citet{nanavati2019follow} and \citet{kastner2022human} represent hybrid approaches that combine learned human interaction or navigation with structured route/follower representations.
Uni-NaVid-FT~\citep{zhang2024uni} represents foundation-model-based navigation fine-tuned on LeadInfra.
We additionally use StarVLA~\citep{ye2026starvla} as a matched direct-trajectory architectural control: it uses the same Qwen3-VL backbone, temporal observations, route/language inputs, prediction horizon, LeadInfra training data, and optimization budget as LeadVLA, but directly predicts the complete waypoint trajectory without Path--Pace--Residual composition.
We also report a Human Expert reference obtained by averaging the results of three human operators evaluated under the same policy-visible observations.

All trainable baselines use the same LeadInfra training split when applicable.
Implementation details and baseline adaptations are provided in Appendix~\ref{app:baseline_details}, and the Human Expert protocol is described in Appendix~\ref{app:human_expert}.

\subsection{Overall Performance on LeadBench}
\label{sec:main_results}

Table~\ref{tab:leadbench_main} shows that LeadVLA achieves the highest overall task success among the evaluated robot policies while maintaining strong route fidelity and follower responsiveness.
The Fiore-style modular method preserves the nominal route particularly well and performs strongly under clean conditions, but achieves lower TRS and SCS, with task success degrading as route interference increases.
In contrast, the matched StarVLA control shows competitive follower responsiveness but degrades more noticeably in route fidelity and task success as execution becomes harder.
These trends are consistent with our motivating observation: retaining the route as an explicit execution reference helps preserve route-level task intent, while learned Pace and Residual provide the flexibility needed for follower- and scene-conditioned adaptation.

The benefit becomes particularly evident in the single-person setting under stronger route interference.
LeadVLA maintains an LSR of 0.50 from Easy to Constrained, while Fiore drops from 0.42 to 0.08 and StarVLA from 0.33 to 0.08.
At the same time, the task remains far from solved: in the Constrained multi-person setting, LeadVLA reaches only 0.04 LSR compared with 0.75 for the Human Expert, highlighting the remaining difficulty of jointly handling route interference, follower interaction, and referent ambiguity.

\begin{table}[t]
\centering
\caption{
\textbf{Overall closed-loop performance on LeadBench.}
Each entry reports Single-person \textbar\ Multi-person
(S \textbar\ M) performance.
For each metric, the top three robot policies are highlighted as
\firstcap{1st}, \secondcap{2nd}, and \thirdcap{3rd}.
Ties share the same rank.
Human Expert is shown in \humanref{gray} as a reference and excluded from ranking.
}
\label{tab:leadbench_main}

\scriptsize
\setlength{\tabcolsep}{0.35pt}
\renewcommand{\arraystretch}{0.96}

\begin{tabularx}{0.995\linewidth}{
@{}l@{\hspace{2.5pt}}
*{4}{Y}
@{\hspace{3.0pt}}
*{4}{Y}
@{\hspace{3.0pt}}
*{4}{Y}
@{}
}
\toprule

&
\multicolumn{4}{c}{\textbf{Core}} &
\multicolumn{4}{c}{\textbf{Easy}} &
\multicolumn{4}{c}{\textbf{Constrained}} \\

\cmidrule(lr){2-5}
\cmidrule(lr){6-9}
\cmidrule(lr){10-13}

\textbf{Method}
& \metrichead{LSR}
& \metrichead{RF}
& \metrichead{TRS}
& \metrichead{SCS}
& \metrichead{LSR}
& \metrichead{RF}
& \metrichead{TRS}
& \metrichead{SCS}
& \metrichead{LSR}
& \metrichead{RF}
& \metrichead{TRS}
& \metrichead{SCS}
\\

\midrule

Fiore et al.
& \sm{\first{.83}}{\second{.38}}
& \sm{\first{.99}}{\first{.94}}
& \sm{\third{.26}}{\second{.24}}
& \sm{.09}{\second{.07}}
& \sm{\second{.42}}{\first{.13}}
& \sm{\first{.96}}{\first{.78}}
& \sm{\third{.27}}{\third{.26}}
& \sm{.05}{.06}
& \sm{.08}{.00}
& \sm{\first{.99}}{\third{.57}}
& \sm{.24}{.25}
& \sm{.09}{\third{.13}}
\\

Nanavati et al.
& \sm{\third{.67}}{\third{.29}}
& \sm{\second{.97}}{\second{.93}}
& \sm{.17}{.14}
& \sm{.07}{.06}
& \sm{\third{.33}}{\third{.08}}
& \sm{\third{.94}}{\third{.71}}
& \sm{.14}{\third{.26}}
& \sm{\third{.13}}{\third{.09}}
& \sm{\second{.17}}{.00}
& \sm{\third{.78}}{\first{.71}}
& \sm{\second{.28}}{.19}
& \sm{\third{.11}}{\third{.13}}
\\

K{\"a}stner et al.
& \sm{.58}{\third{.29}}
& \sm{.81}{.87}
& \sm{.24}{.19}
& \sm{\third{.13}}{.06}
& \sm{\third{.33}}{\third{.08}}
& \sm{.54}{.43}
& \sm{.22}{.14}
& \sm{.05}{.03}
& \sm{\second{.17}}{.00}
& \sm{.47}{.52}
& \sm{.20}{.12}
& \sm{\third{.11}}{.04}
\\

Uni-NaVid-FT
& \sm{.17}{.13}
& \sm{.52}{.43}
& \sm{.22}{\second{.24}}
& \sm{.07}{\second{.07}}
& \sm{.08}{.04}
& \sm{.42}{.39}
& \sm{.19}{\second{.35}}
& \sm{.06}{.06}
& \sm{.00}{.00}
& \sm{.33}{.37}
& \sm{.17}{\second{.29}}
& \sm{.03}{\second{.16}}
\\

StarVLA
& \sm{.42}{.17}
& \sm{.78}{.64}
& \sm{\first{.36}}{\second{.24}}
& \sm{\second{.16}}{\second{.07}}
& \sm{\third{.33}}{.00}
& \sm{.80}{.61}
& \sm{\first{.40}}{\third{.26}}
& \sm{\second{.18}}{\second{.10}}
& \sm{.08}{\first{.04}}
& \sm{.74}{.41}
& \sm{\second{.28}}{\third{.27}}
& \sm{\second{.13}}{.10}
\\

\textbf{LeadVLA}
& \sm{\second{.75}}{\first{.46}}
& \sm{\third{.96}}{\third{.89}}
& \sm{\second{.35}}{\first{.34}}
& \sm{\first{.23}}{\first{.25}}
& \sm{\first{.50}}{\first{.13}}
& \sm{\first{.96}}{\second{.72}}
& \sm{\second{.33}}{\first{.41}}
& \sm{\first{.23}}{\first{.24}}
& \sm{\first{.50}}{\first{.04}}
& \sm{\second{.92}}{\second{.69}}
& \sm{\first{.35}}{\first{.32}}
& \sm{\first{.34}}{\first{.20}}
\\

Human Expert
& \sm{\humanref{1.00}}{\humanref{.96}}
& \sm{\humanref{.99}}{\humanref{.98}}
& \sm{\humanref{.43}}{\humanref{.46}}
& \sm{\humanref{.56}}{\humanref{.51}}
& \sm{\humanref{.83}}{\humanref{.83}}
& \sm{\humanref{.98}}{\humanref{.99}}
& \sm{\humanref{.43}}{\humanref{.41}}
& \sm{\humanref{.49}}{\humanref{.48}}
& \sm{\humanref{.83}}{\humanref{.75}}
& \sm{\humanref{.93}}{\humanref{.91}}
& \sm{\humanref{.43}}{\humanref{.39}}
& \sm{\humanref{.46}}{\humanref{.46}}
\\

\bottomrule
\end{tabularx}

\vspace{-10pt}
\end{table}

\subsection{Design Analysis and Ablations}

We evaluate three key design choices of LeadVLA through controlled studies on
structured route execution,
follower-aware semantic learning and coupling,
and referent-aware co-training.
All variants share the same Qwen3-VL backbone, LeadInfra training data, optimization budget, and LeadBench protocol.
Finer-grained difficulty and response-attribution analyses are provided in Appendix~\ref{app:additional_ablation}, with branch-wise semantic-coupling ablations in Appendix~\ref{app:semantic_coupling}.

\begin{table}[t]
\centering

\begin{minipage}[t]{0.31\linewidth}
\centering

\captionsetup{font=footnotesize,skip=2pt}
\captionof{table}{
\textbf{Structured route execution.}
Results are equally averaged over Core, Easy, and Constrained
single- and multi-person conditions.
}
\label{tab:ablation_structure}

\scriptsize
\renewcommand{\arraystretch}{0.88}
\setlength{\tabcolsep}{1.4pt}

\resizebox{\linewidth}{!}{%
\begin{tabular}{@{}lcccc@{}}
\toprule
\textbf{Formulation}
& LSR$\uparrow$
& RF$\uparrow$
& TRS$\uparrow$
& SCS$\uparrow$ \\
\midrule

Direct Trajectory
& 0.174
& 0.663
& 0.300
& 0.124 \\

Path + Pace
& 0.354
& \textbf{0.988}
& 0.349
& 0.172 \\

Path + Pace + Residual
& \textbf{0.396}
& 0.853
& \textbf{0.352}
& \textbf{0.250} \\

\bottomrule
\end{tabular}%
}

\end{minipage}
\hfill
\begin{minipage}[t]{0.335\linewidth}
\centering

\captionsetup{font=footnotesize,skip=2pt}
\captionof{table}{\textbf{Semantic learning and fusion.} Results are equally averaged over Core, Easy, and Constrained
single- and multi-person conditions.}
\label{tab:ablation_semantic}

\scriptsize
\renewcommand{\arraystretch}{0.88}
\setlength{\tabcolsep}{1.4pt}

\resizebox{\linewidth}{!}{%
\begin{tabular}{@{}lcccc@{}}
\toprule
\textbf{Variant}
& LSR$\uparrow$
& RF$\uparrow$
& TRS$\uparrow$
& SCS$\uparrow$ \\
\midrule

No Semantic
& 0.340
& 0.612
& 0.311
& 0.178 \\

+ Aux. Supervision
& 0.329
& 0.761
& \textbf{0.365}
& 0.180 \\

+ Shared-All Fusion
& 0.333
& 0.817
& 0.357
& 0.228 \\

+ Selective Fusion
& \textbf{0.396}
& \textbf{0.853}
& 0.352
& \textbf{0.250} \\

\bottomrule
\end{tabular}%
}

\end{minipage}
\hfill
\begin{minipage}[t]{0.335\linewidth}
\centering

\captionsetup{font=footnotesize,skip=2pt}
\captionof{table}{\textbf{Referent-aware co-training.} Results are equally averaged over Core, Easy, and Constrained
single- and multi-person conditions.}
\label{tab:ablation_referent}

\scriptsize
\renewcommand{\arraystretch}{0.88}
\setlength{\tabcolsep}{1.4pt}

\resizebox{\linewidth}{!}{%
\begin{tabular}{@{}lcccc@{}}
\toprule
\textbf{Training Design}
& LSR$\uparrow$
& RF$\uparrow$
& TRS$\uparrow$
& SCS$\uparrow$ \\
\midrule

No Intervention
& 0.112
& 0.550
& 0.293
& 0.125 \\

Reassignment Only
& 0.132
& 0.600
& 0.344
& 0.151 \\

Counterfactual Only
& 0.146
& 0.726
& 0.345
& 0.190 \\

\textbf{Full Co-training}
& \textbf{0.396}
& \textbf{0.853}
& \textbf{0.352}
& \textbf{0.250} \\

\bottomrule
\end{tabular}%
}

\end{minipage}

\vspace{-15pt}
\end{table}

\begin{wraptable}{R}{0.39\textwidth}
    \centering
    \captionsetup{font=footnotesize,skip=2pt}
    \caption{
    \textbf{Real-world transfer without policy fine-tuning.}
    Each setting contains 20 closed-loop trials; SR reports successful
    trials out of 20, and SCS is depth-estimated offline.
    }
    \label{tab:real_world}

    \scriptsize
    \setlength{\tabcolsep}{2.0pt}
    \renewcommand{\arraystretch}{0.88}

    \begin{tabular}{@{}lcccc@{}}
    \toprule
    \textbf{Setting}
    & SR$\uparrow$
    & RF$\uparrow$
    & TRS$\uparrow$
    & SCS$\uparrow$ \\
    \midrule

    Core-S
    & 16/20
    & 0.95
    & 0.320
    & 0.210 \\

    Core-M
    & 12/20
    & 0.95
    & 0.305
    & 0.195 \\

    Easy-S
    & 10/20
    & 0.90
    & 0.285
    & 0.165 \\

    Easy-M
    & 4/20
    & 0.80
    & 0.280
    & 0.150 \\

    \bottomrule
    \end{tabular}

\end{wraptable}

\textbf{Does structured route execution matter?}
We compare direct end-to-end trajectory prediction with Path + Pace and the full Path + Pace + Residual formulation.
As shown in Table~\ref{tab:ablation_structure}, introducing Path + Pace increases LSR from 0.174 to 0.354 and RF from 0.663 to 0.988, showing the benefit of retaining the route as an explicit execution reference.
Adding Residual further improves LSR to 0.396 and SCS to 0.250, while reducing RF as the policy gains flexibility for local deviations.
Together, these results support the intended division of roles: Path and Pace stabilize route execution, while Residual provides the local flexibility needed for follower- and scene-conditioned adaptation.

\textbf{How should follower semantics influence action?}
With the structured action formulation fixed, we compare semantic supervision alone, direct fusion of all predicted semantics, and the proposed selective coupling.
Table~\ref{tab:ablation_semantic} shows that auxiliary supervision improves RF and TRS but does not by itself improve LSR, indicating that semantic supervision alone does not necessarily translate into better task completion.
Selective Fusion achieves the highest LSR and SCS among these variants, supporting our design in which semantic objectives broadly shape the shared representation, while direct semantic-to-action coupling is restricted according to the intended control role.
Additional difficulty and branch-wise coupling analyses are provided in Appendices~\ref{app:additional_ablation} and~\ref{app:semantic_coupling}.

\textbf{Does referent-aware co-training improve control attribution?}
Table~\ref{tab:ablation_referent} compares training without referent intervention, with either counterfactual pairs or dynamic reassignment alone, and with both together.
All variants use matched total training scale, sampling budget, and optimization settings.
Either intervention alone yields modest gains in overall task performance, whereas combining both increases LSR from 0.112 to 0.396.
The two forms of supervision address complementary aspects of referent-conditioned control: counterfactual pairs teach \emph{whose} behavior should influence action, while reassignment teaches the policy to update this attribution when the language referent changes.
Consistent with this interpretation, the response-attribution analysis in Appendix~\ref{app:additional_ablation} shows that target-attributed behavior changes increase from 60.0\% without intervention to 86.7\% with full co-training.

\setlength{\intextsep}{3pt}
\setlength{\columnsep}{8pt}

\begin{figure}[!htbp]
    \centering
    \includegraphics[width=0.68\linewidth]{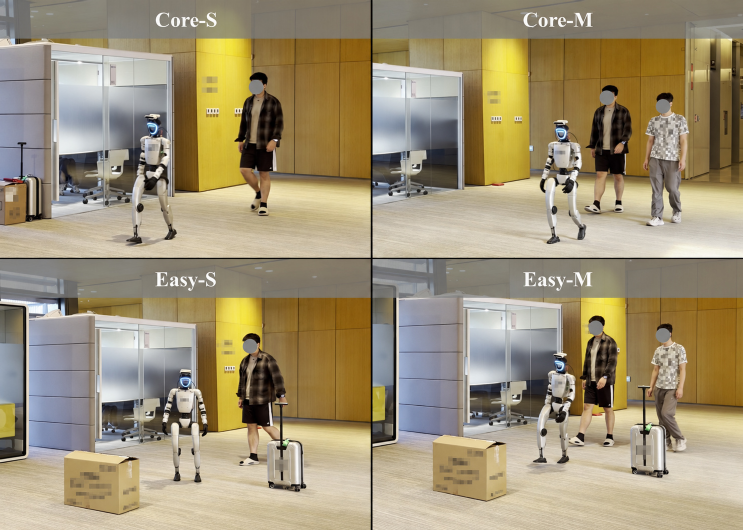}
    \captionsetup{font=footnotesize,skip=2pt}
    \caption{\textbf{Real-world evaluation settings.}
    The physical experiments retain the core interaction structure of LeadBench: Core evaluates clean route execution, Easy adds local route interference, and multi-person settings include a distractor.}
    \label{fig:real_world_settings}
\end{figure}

\subsection{Sim-to-Real Transfer without Real-World Policy Fine-Tuning}
\label{sec:sim2real}

We evaluate whether follower-aware route execution learned entirely in simulation transfers to physical deployment.
LeadVLA is deployed directly on a Unitree G1 humanoid in previously unseen real-world environments, without real-world policy fine-tuning.
The real-world settings retain the same core interaction structure as LeadBench, with simplified route-interference and single-/multi-person conditions.
We evaluate Core-S, Core-M, Easy-S, and Easy-M with 20 closed-loop trials each, using the same front/rear RGB, language, and route interface as in simulation.
At each policy query, LeadVLA predicts a 24-step robot-local trajectory that is executed by the robot's native trajectory-tracking controller without obstacle replanning or modification of the frozen global route.
Inference takes approximately 100--200 ms on an NVIDIA GPU server, while SCS is computed offline from RealSense D455 depth.
Detailed deployment, route construction, controller interfaces, and real-world metric implementation are provided in Appendix~\ref{app:realworld_setup}.

As shown in Fig.~\ref{fig:real_world_settings} and Table~\ref{tab:real_world}, LeadVLA transfers follower-aware route execution across all four real-world settings without real-world policy fine-tuning.
Route fidelity remains high across all settings (0.80--0.95 RF), while the policy retains follower-responsive behavior in both single- and multi-person interactions.
Performance decreases under stronger route interference, indicating that robust real-world leading remains challenging, but these results demonstrate that the route-execution and follower-response behaviors learned from LeadInfra can transfer to previously unseen physical environments.
Additional qualitative analysis and failure cases are provided in Appendix~\ref{app:realworld}; more real-world demonstration videos are available on our \href{https://louiszengCN.github.io/LeadVLA_PAGE/}{project page}.

\section{Conclusion and Discussion}
\label{sec:conclusion}

We studied robot leading as a follower-aware route-execution problem, where a robot must preserve route-level task intent while adapting its execution to a designated follower.
LeadVLA explores a simple structured formulation for this problem: keep the given route as a stable Path reference, while learning through Pace and Residual when and how its execution should adapt to the follower and local scene.
Together with LeadInfra and LeadBench, we provide a unified learning and evaluation framework for studying closed-loop follower-aware robot leading.
Across controlled experiments, LeadVLA achieves the highest task success among the evaluated robot policies on LeadBench and transfers to previously unseen real-world environments without real-world policy fine-tuning.

Robot leading nevertheless remains far from solved.
Performance degrades when strong route interference, multi-person interaction, and referent ambiguity must be handled simultaneously, leaving a substantial gap to human performance.
The current formulation also focuses primarily on adapting locomotion to the follower.
A more natural robot leader could additionally communicate during interaction, for example by turning toward a waiting follower, using gestures, or providing verbal prompts before resuming the task.
Future work can therefore extend follower-aware route execution toward richer, more proactive and communicative leading while improving spatial robustness, referent grounding, and interaction modeling.
We will release LeadVLA checkpoints, LeadInfra, LeadBench, and evaluation tools as an open research stack for continued exploration of richer robot-leading behaviors.

\setlength{\bibsep}{5pt}
\bibliographystyle{unsrtnat}
\bibliography{references}

\appendix
\clearpage
\startcontents[appendix]

\section*{Appendix Table of Contents}

\vspace{0.3em}

\titlecontents{section}
  [0em]
  {\vspace{0.35em}\small\bfseries}
  {\contentslabel{2.3em}}
  {}
  {\titlerule*[0.55pc]{.}\contentspage}

\titlecontents{subsection}
  [1.6em]
  {\small}
  {\contentslabel{3.2em}}
  {}
  {\titlerule*[0.55pc]{.}\contentspage}

\titlecontents{subsubsection}
  [3.5em]
  {\footnotesize}
  {\contentslabel{4.0em}}
  {}
  {\titlerule*[0.55pc]{.}\contentspage}

\setcounter{tocdepth}{2}
\printcontents[appendix]{}{1}{}

\clearpage

\section{Additional Details of LeadVLA}
\label{app:leadvla}

\subsection{Multimodal Inputs and Action Queries}
\label{app:input_queries}

\paragraph{Temporal visual observations.}
LeadVLA uses synchronized front- and rear-view RGB observations to capture both local route-execution context and the evolving state of the follower.
For each view, we use the current frame together with five historical frames approximately uniformly sampled from the preceding 31 simulation frames, covering about 3.1\,s of temporal context at 10\,Hz.
This results in 12 RGB images per policy query.
At the beginning of an episode, unavailable history is filled by repeating the earliest available frame.
All images are resized to $384\times384$ before being processed by the multimodal backbone.

The two views share the same multimodal backbone rather than using camera-specific visual encoders.
We use a fixed input order and explicitly identify the view roles in the instruction: the first six images correspond to the front-view history and current observation, and the following six correspond to the rear view.

\paragraph{Language-conditioned referent, interaction intent, and route.}
The language instruction specifies both \emph{who to lead} and \emph{how to lead}.
In multi-person settings, the designated follower is specified through a natural-language appearance description, while the requested distance and relative formation are expressed in the same instruction.
Referent grounding is learned jointly from appearance-conditioned LeadInfra instructions, target-related semantic supervision, and the controlled referent interventions described in Sec.~\ref{sec:referent_cotraining}; no external person-recognition or ReID supervision is used.

A compact local route hint is serialized directly as robot-local coordinates and inserted into the language prompt, with $x$ denoting the forward direction and $y$ the leftward direction.
The hint contains 10 forward samples from the route reference, spaced by approximately $0.15$\,m in arc length.
This spacing corresponds to the nominal motion increment
$v_{\mathrm{nom}}\Delta t=0.75\,\mathrm{m/s}\times0.2\,\mathrm{s}$,
giving approximately $1.5$\,m of route lookahead.
A representative prompt is:

\begin{quote}
\small
\texttt{You are a guiding robot.}\\
\texttt{The first 6 images are the FRONT history and current view;}\\
\texttt{the next 6 images are the REAR history and current view.}\\
\texttt{Lead the person wearing a teal top and red pants at far distance}\\
\texttt{on the left side.}\\
\texttt{Route waypoints are in robot-local coordinates, in meters:}\\
\texttt{x is forward and y is left.}\\
\texttt{(0.15, 0.00), (0.30, 0.00), (0.45, 0.01), (0.60, 0.01),}\\
\texttt{(0.75, 0.02), (0.90, 0.03), (1.05, 0.04), (1.20, 0.05),}\\
\texttt{(1.35, 0.07), (1.50, 0.09).}\\
\texttt{Predict the next 24 robot waypoints.}
\end{quote}

\paragraph{Dual route representation and shared action queries.}
The route is represented in two complementary forms.
The compact robot-local route hint exposed to the multimodal backbone contains 10 samples and covers approximately $1.5$\,m ahead, allowing local route geometry to be represented jointly with visual observations and language.
In parallel, the 24-point route reference used by the deterministic Path Composer covers approximately $3.6$\,m at the same nominal $0.15$\,m arc-length spacing and is retained outside the multimodal backbone.
Thus, the learned representation receives compact route context, while the longer nominal geometry required for action composition remains explicitly available.

Given the temporal observations, language instruction, and route hint, a single multimodal forward pass produces $H=24$ shared action queries,
\[
Q_t = \{q_{t,1}, \ldots, q_{t,H}\}.
\]
A fixed sequence of 24 action-query tokens is appended to the multimodal prompt, and their final-layer hidden states form the horizon-wise representations.
Pace is predicted from an aggregated representation of these queries, whereas Residual is decoded from the individual horizon queries.
The same representation is additionally supervised for follower-related semantics, as detailed in Sec.~\ref{app:semantic_supervision}.

\subsection{Follower Semantic Supervision}
\label{app:semantic_supervision}

LeadVLA supervises follower-related understanding on the same shared action representation used for control.
Given the horizon-wise action queries
$Q_t=\{q_{t,1},\ldots,q_{t,H}\}$, we compute the pooled representation
\[
\bar q_t = \frac{1}{H}\sum_{k=1}^{H} q_{t,k},
\]
from which six complementary follower-related objectives are predicted, as summarized in Table~\ref{tab:semantic_heads}.

\paragraph{Shared semantic supervision without a state bottleneck.}
The predicted semantics are not assembled into an explicit follower-state vector that is subsequently passed to a separate controller.
Instead, their supervision directly shapes the shared action representation from which Pace and Residual are decoded.
Target localization encourages grounding of the language-designated follower; lateral and distance encode relative geometry; observation captures target presence, loss, and reacquisition; motion describes the follower's movement state; and formation indicates whether the requested spatial relation is being acquired, maintained, or corrected.
The representation therefore retains rich visual and temporal information while being explicitly supervised for follower-relevant structure.

\paragraph{Configurable interaction semantics.}
The supervised spatial relations are defined by the interaction contract used to generate the data, rather than as universal social conventions.
The metric ranges associated with near/far distance and the angular ranges associated with left/center/right formation are configurable in LeadInfra.
LeadVLA therefore learns the interaction relations instantiated by the training contract rather than assuming that a fixed metric distance or bearing is universally ``near,'' ``far,'' ``left,'' or ``right.''
For the experiments in this paper, we use one fixed configuration for training and evaluation; its numerical definition is provided in Sec.~\ref{app:infra_behavior}.

\paragraph{Selective semantic coupling.}
All six objectives shape the shared control representation during training, but only the soft lateral and formation predictions are given an additional direct pathway to Residual.
Their predicted distributions are mapped by the semantic fusion module into a sample-level modulation, which is broadcast to the horizon-wise Residual queries before trajectory decoding.
Using soft distributions preserves prediction uncertainty and keeps the coupling differentiable.

Target localization, distance, observation, and motion have no direct action-fusion pathway and affect action only through the shared representation.
Pace is likewise decoded from the original action queries without explicit semantic fusion.
The output layer of the semantic fusion module is initialized to zero, so the fusion initially contributes no modulation and training begins from the unfused Residual pathway.

\begin{table}[t]
\centering
\small
\caption{
Follower-related semantic supervision used by LeadVLA.
All objectives are predicted from the shared action representation; only lateral and formation distributions are directly fused into the Residual branch.
}
\label{tab:semantic_heads}
\begin{tabular}{lll}
\toprule
\textbf{Objective} & \textbf{Prediction target} & \textbf{Role} \\
\midrule
Target localization
& target bbox
& referent grounding \\

Lateral relation
& left / center / right
& relative spatial relation \\

Distance relation
& too close / in range / too far
& relation to requested gap \\

Observation state
& tracked / missing / lost / reacquiring
& temporal target presence \\

Motion state
& moving / slowing / stopped
& follower motion \\

Formation state
& acquiring / maintaining / correcting
& local interaction state \\
\bottomrule
\end{tabular}
\end{table}

\subsection{Path Composer}
\label{app:path_composer}

The Path Composer is deterministic and contains no learned parameters.
It converts the scalar Pace prediction into a smooth horizon-wise execution profile, queries the nominal route according to the resulting progress, and composes the learned Residual to obtain the final trajectory.

\paragraph{Rate-limited Pace profile.}
At time $t$, the Pace branch predicts a target progress scale
$\hat a_t\in[0,1]$.
Rather than applying this value instantaneously over the full prediction horizon, the Composer initializes from the Pace executed in the previous control cycle,
$a^{\mathrm{exec}}_{t-1}$, and generates a rate-limited future profile
$\hat a_{t,1:H}$.

Let $v_{\mathrm{nom}}$ denote the nominal leading speed. We define
\[
v_{\mathrm{target}}=\hat a_t v_{\mathrm{nom}},
\qquad
v_0=a^{\mathrm{exec}}_{t-1}v_{\mathrm{nom}},
\]
and recursively update
\[
v_k =
v_{k-1}
+
\mathrm{clip}
\left(
v_{\mathrm{target}}-v_{k-1},
-a_{\mathrm{dec}}\Delta t,
a_{\mathrm{acc}}\Delta t
\right),
\]
with
\[
\hat a_{t,k}
=
\mathrm{clip}
\left(
\frac{v_k}{v_{\mathrm{nom}}},
0,1
\right).
\]
This converts the learned target Pace into smooth slowdown, waiting, and resumption without introducing discrete hand-designed execution states.

\paragraph{Arc-length composition.}
Let $P_{t,0}$ denote the current route origin at relative progress $s_{t,0}=0$, followed by the $H$ future route knots
$\{(P_{t,k},s_{t,k})\}_{k=1}^{H}$.
With
$\Delta s_{t,k}=s_{t,k}-s_{t,k-1}$,
the Pace-conditioned queried progress is
\[
\hat s_{t,k}
=
\sum_{j=1}^{k}
\Delta s_{t,j}\hat a_{t,j}.
\]
The nominal route is then interpolated at this progress,
\[
\tilde P_{t,k}=P_t(\hat s_{t,k}).
\]
For position, the Composer uses piecewise-linear interpolation in robot-local $x$--$y$ coordinates.
Heading is first unwrapped, linearly interpolated, and then wrapped back to $[-\pi,\pi)$.
Thus, Pace changes how far the robot advances along the nominal route without directly deforming its geometry.

\paragraph{Residual composition.}
The learned horizon-wise Residual is composed with the pace-conditioned reference:
\[
\hat \tau_{t,k}
=
\tilde P_{t,k}
\oplus
\hat r_{t,k},
\]
where $\oplus$ denotes addition in robot-local position and wrapped addition in heading.
Residual is not restricted to lateral follower correction; it captures local adaptation not explained by route progress alone, including formation correction, scene-dependent deviation, and recovery toward the nominal route.

The corresponding Pace and Residual training targets are defined in Sec.~\ref{app:training}.

\begin{table}[t]
\centering
\small
\caption{Fixed parameters of the Path Composer used in our experiments.}
\label{tab:path_composer_config}
\begin{tabular}{lc}
\toprule
\textbf{Parameter} & \textbf{Value} \\
\midrule
Prediction horizon $H$ & 24 \\
Trajectory interval $\Delta t$ & $0.2$ s \\
Nominal speed $v_{\mathrm{nom}}$ & $0.75$ m/s \\
Acceleration limit $a_{\mathrm{acc}}$ & $0.5$ m/s$^2$ \\
Deceleration limit $a_{\mathrm{dec}}$ & $0.6$ m/s$^2$ \\
\bottomrule
\end{tabular}
\end{table}

\subsection{Training Objective and Configuration}
\label{app:training}

\paragraph{Action-target construction.}
For each training frame, the causal expert provides a future robot-local waypoint trajectory
$\tau^{\mathrm{gt}}_{t,1:H}$
together with a scalar target Pace
$a^{\mathrm{gt}}_t\in[0,1]$.
The Pace target is generated online from the realized robot--follower interaction rather than recovered post hoc from the expert trajectory.

To construct the Residual target, we pass $a^{\mathrm{gt}}_t$ through the same deterministic, rate-limited Path Composer described in Sec.~\ref{app:path_composer}, initialized from the previously executed Pace.
This produces the target pace-conditioned route reference
$\tilde P^{\mathrm{gt}}_{t,1:H}$.
The Residual target is then defined relative to this reference:
\[
r^{\mathrm{gt},xy}_{t,k}
=
\tau^{\mathrm{gt},xy}_{t,k}
-
\tilde P^{\mathrm{gt},xy}_{t,k},
\]
\[
r^{\mathrm{gt},\psi}_{t,k}
=
\mathrm{wrap}
\left(
\tau^{\mathrm{gt},\psi}_{t,k}
-
\tilde P^{\mathrm{gt},\psi}_{t,k}
\right).
\]
Thus, once the nominal route and expert Pace are fixed, the decomposition into pace-conditioned Path and Residual is deterministic.
For frames in which follower-reaction supervision is not defined, the Pace loss is masked and the Composer uses unit Pace, leaving setup or formation-specific motion to Residual.

\paragraph{Joint action and semantic objectives.}
Action-related regression uses Smooth L1 loss with task-specific validity masks.
Let $\mathcal{M}_{\mathrm{traj}}$ denote valid future trajectory elements and
$\mathcal{M}_{\mathrm{react}}$ denote frames with valid Pace supervision.
The three primary action losses are
\[
\mathcal{L}_{\mathrm{wp}}
=
\operatorname{mean}_{\mathcal{M}_{\mathrm{traj}}}
\mathrm{SmoothL1}
\left(
\hat\tau_{t,k},
\tau^{\mathrm{gt}}_{t,k}
\right),
\]
\[
\mathcal{L}_{\mathrm{pace}}
=
\operatorname{mean}_{\mathcal{M}_{\mathrm{react}}}
\mathrm{SmoothL1}
\left(
\hat a_t,
a^{\mathrm{gt}}_t
\right),
\]
and
\[
\mathcal{L}_{\mathrm{res}}
=
\operatorname{mean}_{\mathcal{M}_{\mathrm{traj}}}
\mathrm{SmoothL1}
\left(
\hat r_{t,k},
r^{\mathrm{gt}}_{t,k}
\right).
\]

We additionally regularize the magnitude and within-horizon variation of the predicted Residual:
\[
\mathcal{L}_{\mathrm{norm}}
=
\operatorname{mean}_{\mathcal{M}_{\mathrm{traj}}}
\left|
\hat r_{t,k}
\right|,
\]
\[
\mathcal{L}_{\mathrm{smooth}}
=
\operatorname{mean}_{\mathcal{M}_{\mathrm{pair}}}
\left|
\hat r_{t,k}
-
\hat r_{t,k-1}
\right|,
\]
where $\mathcal{M}_{\mathrm{pair}}$ contains adjacent future steps for which both waypoints are valid.
The first term discourages Residual from unnecessarily replacing the nominal route reference, while the second encourages a smooth 24-step correction trajectory.
The action objective is
\[
\mathcal{L}_{\mathrm{act}}
=
\mathcal{L}_{\mathrm{wp}}
+
\mathcal{L}_{\mathrm{pace}}
+
\mathcal{L}_{\mathrm{res}}
+
0.001\,\mathcal{L}_{\mathrm{norm}}
+
0.02\,\mathcal{L}_{\mathrm{smooth}}.
\]

The semantic objective supervises the six follower-related predictions described in Sec.~\ref{app:semantic_supervision}:
\[
\mathcal{L}_{\mathrm{sem}}
=
\mathcal{L}_{\mathrm{bbox}}
+
0.5
\left(
\mathcal{L}_{\mathrm{lat}}
+
\mathcal{L}_{\mathrm{dist}}
+
\mathcal{L}_{\mathrm{obs}}
+
\mathcal{L}_{\mathrm{motion}}
+
\mathcal{L}_{\mathrm{form}}
\right).
\]
Target localization uses Smooth L1 regression on normalized bounding boxes, whereas the five semantic classification objectives use cross-entropy loss.
The final objective is
\[
\mathcal{L}
=
\mathcal{L}_{\mathrm{act}}
+
0.25\,\mathcal{L}_{\mathrm{sem}}.
\]

Task-specific validity masks exclude undefined supervision.
Target localization is optimized only when a valid target box is available; each semantic objective is applied only when its corresponding label is defined; and padded future waypoints near episode termination are excluded from waypoint, Residual, and smoothness losses.

\paragraph{Two-stage scale-up training.}
We use a two-stage scale-up training protocol.
In Stage~1, the full LeadVLA architecture is initialized from Qwen3-VL-2B-Instruct~\cite{yang2025qwen3} and trained on an approximately 200K-sample subset drawn from the LeadInfra training distribution.
This stage provides a stable initialization before scaling to the full training corpus.

In Stage~2, we load the selected Stage~1 model weights, reinitialize the optimizer and learning-rate scheduler, and continue training on the complete LeadInfra corpus containing approximately 4M training samples.

The two stages use the same model architecture, objectives, supervision definitions, and sampling strategy, differing primarily in training-data scale.
The Qwen3-VL vision tower remains frozen throughout both stages, while the remaining multimodal backbone, Pace and Residual branches, auxiliary semantic heads, and Selective Semantic Fusion module are jointly optimized.

All controlled LeadVLA variants used in the ablation studies follow the same two-stage protocol with matched data scale, sampling strategy, optimizer configuration, and checkpoint-selection procedure.
Thus, differences among these variants reflect the corresponding architectural or training intervention rather than unequal training resources.

\begin{table}[t]
\centering
\small
\caption{
Two-stage scale-up training used for LeadVLA and all controlled ablation variants.
}
\label{tab:training_stages}
\begin{tabular}{lcc}
\toprule
\textbf{Configuration} & \textbf{Stage 1} & \textbf{Stage 2} \\
\midrule
Training-data scale
& $\sim$200K
& $\sim$4M \\

Hardware
& 16$\times$ NVIDIA H20 96GB
& 16$\times$ NVIDIA H20 96GB \\

Wall-clock time
& 2.26 h
& 19.66 h \\

GPU-hours
& 36.16
& 314.63 \\
\bottomrule
\end{tabular}
\end{table}

\paragraph{Optimization and compute.}
We use AdamW with
$\beta_1=0.9$, $\beta_2=0.95$, $\epsilon=10^{-8}$,
and weight decay $10^{-8}$.
The non-frozen Qwen3-VL backbone uses a learning rate of $2\times10^{-5}$,
the Residual branch $1\times10^{-5}$,
and the Pace head, semantic heads, and fusion module $1\times10^{-4}$.
We use a constant learning-rate schedule with linear warmup and clip gradients to a global norm of 1.0.

Training uses BF16 mixed precision, DeepSpeed ZeRO-2, and gradient checkpointing.
For each stage, we select the checkpoint with the lowest validation total loss.
The complete two-stage training takes approximately 21.92 hours on 16 NVIDIA H20 GPUs, corresponding to 350.79 GPU-hours.

\section{LeadInfra: Interaction Supervision for Robot Leading}
\label{app:leadinfra}

\paragraph{Positioning among existing resources.}
Table~\ref{tab:leadinfra_comparison} compares LeadInfra with representative guide-related resources and human-aware navigation simulators.
We focus on capabilities relevant to constructing follower-aware route-execution data, including route geometry, human behavior, human--robot interaction, embodied assets, scene diversity, and closed-loop execution.
The comparison reflects capabilities directly exposed by each released resource or framework, rather than functionality that could be added through substantial task-specific engineering.

\begin{table}[t]
\centering
\scriptsize
\setlength{\tabcolsep}{1.4pt}
\renewcommand{\arraystretch}{1.12}

\caption{
\textbf{Comparison with representative guide-related resources and human-aware navigation simulators.}
$\checkmark$ denotes direct and substantial support,
$\triangle$ denotes partial, limited, or host-dependent support,
and -- denotes that the capability is not directly provided.
Human--robot interaction refers to explicit interaction beyond passive human presence.
}
\label{tab:leadinfra_comparison}

\resizebox{\linewidth}{!}{
\begin{tabular}{
@{}l
l
c
c
c
c
c
c
c
c@{}
}
\toprule

\textbf{Resource}
&
\textbf{Base}
&
\shortstack{\textbf{Diverse}\\\textbf{route}\\\textbf{geometry}}
&
\shortstack{\textbf{Diverse}\\\textbf{human}\\\textbf{behavior}}
&
\shortstack{\textbf{Human--robot}\\\textbf{interaction}}
&
\shortstack{\textbf{Diverse}\\\textbf{actor}\\\textbf{appearance}}
&
\shortstack{\textbf{Diverse}\\\textbf{sensors}}
&
\shortstack{\textbf{Diverse}\\\textbf{robot}\\\textbf{platforms}}
&
\shortstack{\textbf{Diverse}\\\textbf{scenes}}
&
\shortstack{\textbf{Closed}\\\textbf{loop}}
\\

\midrule

\multicolumn{10}{@{}l}{\textit{Guide-related data and benchmarks}} \\[1pt]

GuideData~\cite{hwang2026guidenav}
&
Real-world video
&
$\triangle$
&
$\checkmark$
&
$\checkmark$
&
$\triangle$
&
--
&
--
&
$\triangle$
&
--
\\

SAIT / SA-Bench~\cite{han2025space}
&
Real-world images
&
$\triangle$
&
--
&
--
&
--
&
--
&
--
&
$\checkmark$
&
--
\\

\midrule

\multicolumn{10}{@{}l}{\textit{Human-aware navigation simulators and frameworks}} \\[1pt]

CARLA~\cite{dosovitskiy2017carla,zeng2026carla}
&
UE
&
$\checkmark$
&
$\triangle$
&
$\triangle$
&
$\checkmark$
&
$\checkmark$
&
$\triangle$
&
$\triangle$
&
$\checkmark$
\\

SEAN 2.0~\cite{tsoi2022sean}
&
Unity
&
$\triangle$
&
$\checkmark$
&
$\checkmark$
&
$\triangle$
&
$\triangle$
&
$\triangle$
&
$\triangle$
&
$\checkmark$
\\

HuNavSim~\cite{perez2023hunavsim}
&
ROS2 + host simulator
&
$\triangle$
&
$\checkmark$
&
$\checkmark$
&
$\triangle$
&
$\triangle$
&
$\triangle$
&
$\triangle$
&
$\checkmark$
\\

Habitat 3.0~\cite{puig2024habitat}
&
Habitat-Sim
&
$\triangle$
&
$\checkmark$
&
$\checkmark$
&
$\triangle$
&
$\checkmark$
&
$\checkmark$
&
$\checkmark$
&
$\checkmark$
\\

NavIsaacLab~\cite{xia2026navisaaclab}
&
Isaac Lab
&
$\triangle$
&
$\checkmark$
&
$\triangle$
&
$\triangle$
&
$\checkmark$
&
$\checkmark$
&
$\checkmark$
&
$\checkmark$
\\

\midrule

\textbf{LeadInfra}
&
\textbf{UE}
&
\textbf{$\checkmark$}
&
\textbf{$\checkmark$}
&
\textbf{$\checkmark$}
&
\textbf{$\checkmark$}
&
\textbf{$\checkmark$}
&
\textbf{$\checkmark$}
&
\textbf{$\checkmark$}
&
\textbf{$\checkmark$}
\\

\bottomrule
\end{tabular}
}
\end{table}

\subsection{LeadInfra Assets and Environment Overview}
\label{app:infra_assets}

\begin{figure*}[t]
    \centering
    \includegraphics[width=\textwidth]{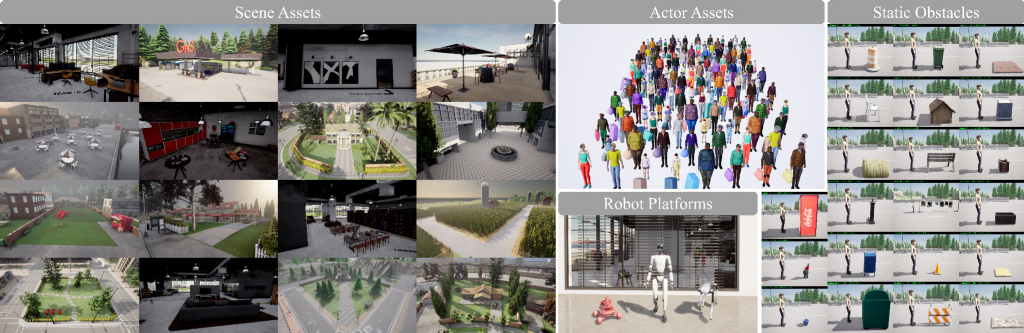}
    \caption{
    \textbf{Representative assets in LeadInfra.}
    LeadInfra combines diverse indoor and outdoor environments with multiple robot embodiments, compositional human appearances, sensing modalities, and physical scene assets.
    These assets support the route, interaction, and scene variations introduced in the following sections.
    }
    \label{fig:leadinfra_assets}
\end{figure*}

LeadInfra provides the environment and asset substrate for constructing diverse robot-leading episodes, including indoor and outdoor environments, robot embodiments, human appearances, sensing modalities, and reusable scene assets.
The major asset groups are summarized in Table~\ref{tab:leadinfra_assets} and visualized in Fig.~\ref{fig:leadinfra_assets}.

\paragraph{Environments and maps.}
LeadInfra covers approximately $65{,}198~\mathrm{m}^2$ of navigable indoor and outdoor space, including parks, playgrounds, residential areas, offices, exhibition spaces, restaurants, and parking areas.
Each environment is accompanied by a metric GridMap at $0.5~\mathrm{m/cell}$ resolution, providing a common spatial representation for route generation and scene construction.

\paragraph{Embodied and human assets.}
The infrastructure supports three representative robot embodiments: a Unitree G1 humanoid, a Scout Mini wheeled robot, and a quadruped robot-dog platform.
For followers and distractors, LeadInfra provides 756 compatibility-validated human appearance configurations composed from variations in body, clothing, accessories, and body scale.
This appearance diversity supports language-conditioned target designation in multi-person interactions.

\paragraph{Sensing and physical scene assets.}
LeadInfra provides 25 sensor blueprints spanning camera, LiDAR, and state-sensing modalities, together with 142 reusable static objects for constructing scene interference, occlusion, and local route variation.
These numbers describe the capability of the infrastructure rather than the observation space of LeadVLA: the policy studied in this work uses only the front/rear RGB observations and task inputs described in Sec.~\ref{app:input_queries}, while additional sensing modalities are used only for data construction, supervision, or evaluation where required.

\begin{table}[t]
\centering
\small
\caption{
\textbf{Major asset groups in LeadInfra.}
}
\label{tab:leadinfra_assets}
\begin{tabular}{lll}
\toprule
\textbf{Asset group} & \textbf{Scale} & \textbf{Representative coverage} \\
\midrule

Scenes / GridMaps
& $65{,}198~\mathrm{m}^2$
& indoor + outdoor; $0.5~\mathrm{m/cell}$ \\

Robot embodiments
& 3
& humanoid / wheeled / quadruped \\

Human appearances
& 756
& body / clothing / accessories / scale \\

Sensor blueprints
& 25
& camera / LiDAR / state sensing \\

Static objects
& 142
& everyday and structural assets \\

\bottomrule
\end{tabular}
\end{table}

\subsection{Route Generation and Scene Variation}
\label{app:infra_route}

\paragraph{Route generation.}
LeadInfra generates nominal routes from the metric navigability representation of each environment.
Starting from the $0.5~\mathrm{m/cell}$ GridMap introduced in Sec.~\ref{app:infra_assets},
the route generator samples feasible anchors in navigable space, connects them through collision-free paths, and smooths the resulting trajectories into arc-length-parameterized routes $P(s)$.
Routes are filtered for traversability, clearance, continuity, and curvature, while geometrically redundant routes are removed to reduce repeated sampling of the same spatial corridor.

The resulting collection contains approximately 1K nominal routes spanning six representative geometry families:
straight, single-turn, S-shaped, multi-turn, U-turn, and loop trajectories.
Route geometry is varied independently from follower interaction programs, allowing similar events such as lagging, stopping, or reacquisition to occur across different route structures.
This discourages the policy from associating a particular follower response with a fixed trajectory pattern and instead exposes follower-aware adaptation across diverse route executions.

\paragraph{Nominal route and local scene variation.}
Each episode is anchored by a nominal route that remains fixed throughout execution.
LeadInfra then varies the local physical conditions under which this route must be executed.
In selected episodes, static objects or pedestrians create local route interference, requiring temporary deviation from the nominal route followed by recovery toward it.

These cases couple local route adaptation with follower-aware interaction.
For example, a temporary deviation may alter robot--follower distance and bearing, change the rear-view observation, or briefly occlude the designated follower.
The policy must therefore adapt locally while preserving both the nominal route reference and appropriate follower-aware behavior.
These variations are intended to study route recovery under interaction rather than to define a general-purpose obstacle-avoidance task.

\paragraph{Route information boundary.}
The learned policy always receives the original nominal route as its route input.
For data generation, LeadInfra may internally construct episode-specific collision-free execution references for the causal expert and the physical motion of human actors.
These privileged execution references are never exposed to the learned policy.
The policy therefore receives neither an expert detour nor an obstacle-conditioned replacement route or local-planning output.

Any local deviation must consequently be inferred from policy-visible observations while the nominal route remains the persistent task-level spatial reference.
This information boundary directly matches the structured formulation of LeadVLA:
Path preserves the nominal route, while learned Pace and Residual determine how that route is executed under follower and local scene variation.

\subsection{Follower Interaction Programs}
\label{app:infra_behavior}

LeadInfra structures follower interaction along three complementary dimensions:
\emph{motion}, \emph{presence}, and \emph{requested formation}.
These factors are configurable and can be composed with different routes and scenes, subject to physical and semantic compatibility.
This allows similar follower events to occur under diverse execution conditions rather than being tied to a particular route or scene.

\paragraph{Follower motion programs.}
We define eight representative temporal motion programs covering stable tracking, different degrees of lagging and slowdown, hesitation, stopping, recovery, and failure to initiate.
Each program specifies a behavior pattern with randomized timing and motion parameters rather than a fixed follower trajectory.
The programs are summarized in Table~\ref{tab:follower_motion_programs}.

\begin{table}[t]
\centering
\footnotesize
\setlength{\tabcolsep}{3pt}
\renewcommand{\arraystretch}{1.08}
\caption{
\textbf{Follower-motion programs in LeadInfra.}
The final column describes the intended interaction consequence;
the hidden program state itself is never directly exposed to the robot expert or learned policy.
}
\label{tab:follower_motion_programs}
\begin{tabularx}{\linewidth}{@{}lXX@{}}
\toprule
\textbf{Program} &
\textbf{Follower behavior} &
\textbf{Typical leading response} \\
\midrule

Steady tracking
& maintains nominal following
& continue leading \\

Mild lag
& gradually falls slightly behind
& reduce Pace \\

Persistent lag
& sustains a large separation
& strong slowdown / wait \\

Short pause
& briefly stops and resumes
& limited slowdown \\

Gradual slowdown
& continuously reduces walking speed
& smooth Pace adaptation \\

Multi-hesitate
& repeatedly alternates lag and recovery
& repeated Pace adaptation \\

Stop--recover
& stops for an extended period, then catches up
& wait and resume \\

Start failure
& does not begin following after task onset
& stop and wait \\

\bottomrule
\end{tabularx}
\end{table}

The program labels control only the simulated follower behavior and are not directly mapped to robot actions.
Robot supervision is generated from the realized physical consequences of these programs, such as sustained separation or loss of target observability, as detailed in Sec.~\ref{app:infra_expert}.

\paragraph{Presence and reacquisition.}
Motion variation is complemented by explicit target-presence conditions.
LeadInfra distinguishes a follower that slows or falls behind while remaining observable from one that becomes temporarily or persistently unobservable.
We use five representative conditions, summarized in Table~\ref{tab:follower_presence_programs}.

\begin{table}[t]
\centering
\footnotesize
\setlength{\tabcolsep}{3pt}
\renewcommand{\arraystretch}{1.08}
\caption{
\textbf{Target-presence conditions in LeadInfra.}
}
\label{tab:follower_presence_programs}
\begin{tabularx}{\linewidth}{@{}lX@{}}
\toprule
\textbf{Condition} & \textbf{Description} \\
\midrule

Normal visible
& target remains continuously observable \\

Stationary fall-behind
& target stops and progressively increases separation \\

Temporary occlusion
& target is briefly hidden by local scene geometry \\

Target absent, distractor visible
& designated target leaves view while another person remains visible \\

Empty rear / reacquisition
& rear view becomes empty and the target later re-enters \\

\bottomrule
\end{tabularx}
\end{table}

These conditions introduce temporal ambiguity that cannot be resolved from a single observation.
For example, temporary occlusion during local route adaptation should not necessarily trigger immediate waiting, whereas sustained loss of the designated follower should eventually do so.
Likewise, continued visibility of a distractor must not substitute for the missing target.

\paragraph{Configurable interaction contract.}
LeadInfra also varies the requested robot--follower relation through language-conditioned distance and formation specifications.
These relations are configurable interaction requirements rather than universal social-distance conventions.
For the experiments in this paper, we use the fixed configuration in Table~\ref{tab:interaction_contract}.
Combining two distance relations with three relative formations yields six language-conditioned formation settings.

\begin{table}[t]
\centering
\small
\caption{
\textbf{Interaction-contract configuration used for training and evaluation.}
}
\label{tab:interaction_contract}
\begin{tabular}{lc}
\toprule
\textbf{Relation} & \textbf{Configured range} \\
\midrule
Near distance & $1.10$--$1.60$ m \\
Far distance & $2.60$--$4.00$ m \\
Center bearing & $[-5^\circ,+5^\circ]$ \\
Left bearing & $[+15^\circ,+25^\circ]$ \\
Right bearing & $[-25^\circ,-15^\circ]$ \\
\bottomrule
\end{tabular}
\end{table}

These interaction-contract definitions are used to construct follower-related semantic supervision and the corresponding benchmark conditions.
Thus, LeadVLA learns the interaction contract instantiated by the data rather than assuming a universally correct metric distance or bearing.

\subsection{Referent-Aware Interaction}
\label{app:infra_referent}

\begin{figure*}[t]
    \centering
    \includegraphics[width=\textwidth]{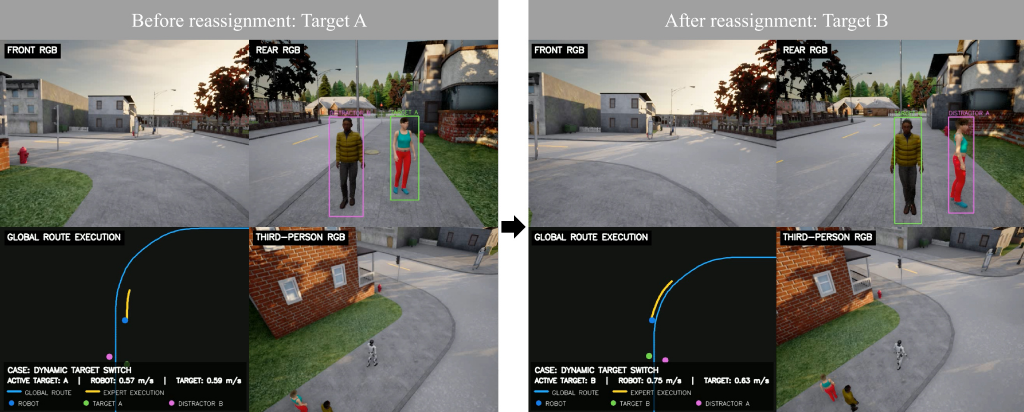}
    \caption{
    \textbf{Dynamic referent reassignment in LeadInfra.}
    The designated follower changes from Target A to Target B while both people remain in the scene and the nominal route is unchanged.
    After reassignment, subsequent robot behavior should depend on the newly designated follower.
    Colored boxes and identity labels are shown only for visualization and are never provided to the policy.
    }
    \label{fig:dynamic_referent_reassignment}
\end{figure*}

LeadInfra constructs multi-person interactions in which one person is designated as the follower through language while other nearby people act as distractors.
This allows follower-conditioned control to be supervised with respect to the current language referent rather than unrelated distractor behavior in the scene.

\paragraph{Appearance-conditioned target designation.}
In multi-person episodes, the policy-visible target designation is provided through a natural-language appearance description together with the requested distance and formation.
No target ID, ground-truth bounding box, person trajectory, or simulator identity is exposed to the policy.
Target and distractor appearances are selected to remain distinguishable under their corresponding language descriptions, using the human appearance assets introduced in Sec.~\ref{app:infra_assets}.

\paragraph{Counterfactual referent pairs.}
We construct matched episode pairs that preserve the environment, nominal route, human actors, initial configuration, sensing conditions, and human behavior, while changing which person is designated by the language instruction.
For example, one person may execute a stop--recover behavior while another continues steady tracking.
If the stopping person is the target, the resulting interaction should induce slowing or waiting; if the other person is designated instead, the same stop becomes a distractor event and should not modulate the robot response.

Because the two cases differ primarily in the referent, these pairs provide controlled supervision for target-conditioned action attribution:
\[
\text{target behavior} \rightarrow \text{action modulation},
\qquad
\text{distractor behavior} \rightarrow \text{ignore}.
\]

\paragraph{Dynamic target reassignment.}
Selected episodes additionally change the designated follower once during execution.
When the new target satisfies predefined visibility and formation-compatibility conditions, the target and distractor roles are atomically reassigned and the language instruction is updated to describe the new follower.
No explicit ``switch'' state or target identifier is provided to the policy; it simply receives the updated ordinary leading instruction.

The previously designated follower remains in the scene after reassignment and continues moving, but should no longer determine the robot response.
Conversely, subsequent behavior of the newly designated follower should now influence execution.
This complements the counterfactual pairs by requiring referent-conditioned control to update online rather than remain attached to a persistent person identity.

Together, counterfactual referent pairs and dynamic reassignment provide the referent-aware co-training supervision described in Sec.~\ref{sec:referent_cotraining}.

\subsection{Causal Expert Supervision}
\label{app:infra_expert}

LeadInfra uses scripted follower programs to generate controlled interaction events, but these hidden programs are never mapped directly to robot actions.
Instead, the programs first produce physical follower behavior in simulation, and the expert reacts only to interaction evidence that has already been realized:
\[
\text{hidden follower program}
\rightarrow
\text{realized physical interaction}
\rightarrow
\text{expert robot action}.
\]

The expert may use simulator-derived information describing the current realized interaction, including robot--follower geometry, target observability, robot state, and local scene feasibility.
Latent behavior labels, future schedules, and upcoming interaction transitions are excluded from the reaction logic.
Robot trajectories are generated online in closed loop from the current realized state rather than replayed from precomputed action scripts.

\paragraph{Distance-driven Pace supervision.}
Follower separation is measured along the current execution route rather than by Euclidean distance.
Let
\[
g_t=s_t^{\mathrm{robot}}-s_t^{\mathrm{target}}
\]
denote the robot--target arc-length separation.
At the transition from setup to leading, the realized separation is recorded as a baseline $g_0$.
After dynamic target reassignment, the baseline is reset according to the newly requested formation.

To suppress short-term motion noise, the expert maintains an exponentially smoothed separation,
\[
\bar g_t
=
\bar g_{t-1}
+
\alpha
\left(
g_t-\bar g_{t-1}
\right),
\qquad
\alpha=1-\exp(-\Delta t/0.5),
\]
corresponding to a $0.5$\,s time constant.
We define the realized separation drift as
\[
d_t=\bar g_t-g_0.
\]

The expert begins reducing Pace only when
$d_t>0.25$\,m persists for at least $0.5$\,s.
For sustained drift, the target Pace scale is
\[
z_t
=
\operatorname{clip}
\left(
\frac{d_t-0.25}{1.25-0.25},
0,1
\right),
\qquad
a_t^{\mathrm{gt}}
=
1-z_t^2(3-2z_t).
\]
Thus, the target remains at full Pace for small separation changes, decreases smoothly as the follower falls behind, and reaches zero when the drift reaches $1.25$\,m.
The resulting expert motion is rate-limited with acceleration and deceleration bounds of
$0.5$ and $0.6$\,m/s$^2$, respectively, at the nominal leading speed of $0.75$\,m/s.

\paragraph{Target loss and reacquisition.}
Target observability provides a second source of causal interaction evidence.
When the designated follower first becomes unobservable, the expert enters a conservative temporary-missing regime rather than stopping immediately.
During this period, the Pace target is capped at $0.65$.
If the target remains continuously missing for $0.8$\,s or the robot travels $0.6$\,m without observing the target, whichever occurs first, the expert transitions to waiting and reduces the target Pace to zero through a smooth braking transition.

When the same target reappears, it must remain continuously observable for $0.5$\,s before normal leading resumes.
A new disappearance during this confirmation period returns the expert to waiting.
Resumption is likewise rate-limited rather than applied as an instantaneous speed change.

Temporary occlusion caused by static scene geometry is treated separately from persistent target loss.
Such occlusions retain their visibility and occlusion supervision, but short static occlusions do not directly trigger the persistent-loss stopping logic in the final data collection.
This distinction exposes the policy to cases in which the follower is briefly unobservable during local route adaptation without teaching it to stop at every transient occlusion.

\paragraph{Causal supervision boundary.}
These rules depend only on interaction evidence available after it has physically occurred.
For example, a scripted stop does not immediately reduce the expert Pace; the response emerges only after the realized robot--follower separation exceeds the reaction threshold.
Similarly, target loss affects the expert only after sustained absence or blind travel is observed.
The resulting supervision therefore preserves precise simulator-based control while avoiding shortcut mappings from hidden follower programs or future interaction schedules to robot actions.

\subsection{Data Scale and Composition}
\label{app:infra_composition}

\begin{figure*}[t]
    \centering
    \includegraphics[width=\textwidth]{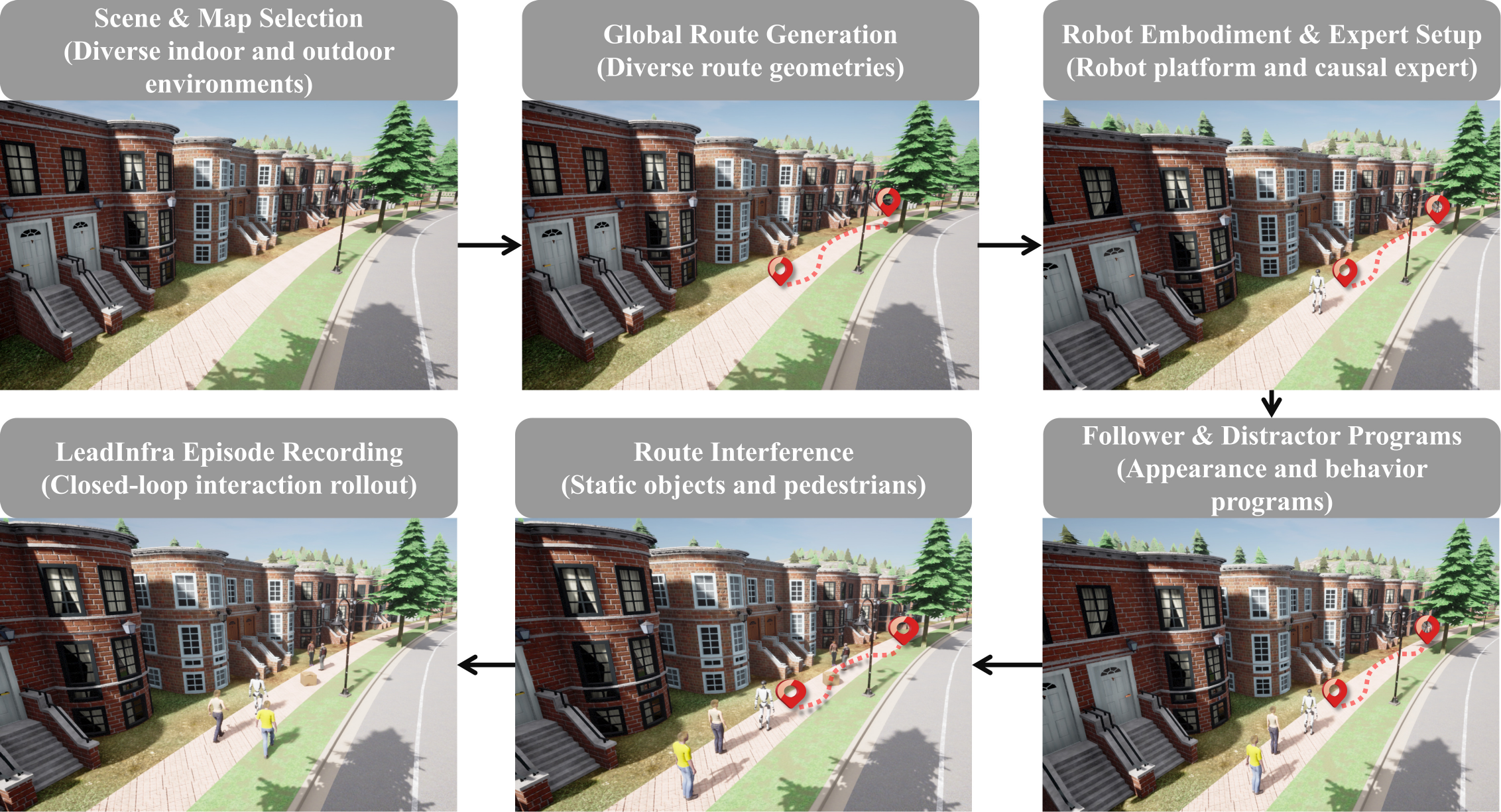}
    \caption{
    \textbf{LeadInfra scene composition and episode generation.}
    Starting from diverse environments (Sec.~\ref{app:infra_assets}), LeadInfra samples a nominal route (Sec.~\ref{app:infra_route}),
    instantiates robot and human actors, composes follower-interaction programs and local scene variation
    (Secs.~\ref{app:infra_behavior} and~\ref{app:infra_referent}), and executes the resulting task in closed loop.
    A causal expert (Sec.~\ref{app:infra_expert}) generates robot supervision from realized interaction evidence rather than directly from hidden follower programs.
    }
    \label{fig:leadinfra_composition}
\end{figure*}

The current LeadInfra corpus contains approximately 4M training samples collected over roughly 1K nominal routes.
Unlike short isolated clips, episodes are temporally extended closed-loop interactions that may contain multiple follower events and repeated transitions among leading, slowing, waiting, and recovery.
Table~\ref{tab:leadinfra_composition} summarizes the scale and major variation dimensions of the current training corpus.

\begin{table}[t]
\centering
\small
\caption{
\textbf{Scale and composition of the current LeadInfra training corpus.}
Route length denotes the arc length of the nominal route used to construct each episode.
}
\label{tab:leadinfra_composition}
\begin{tabular}{lc}
\toprule
\textbf{Dimension} & \textbf{Coverage} \\
\midrule

Training samples
& $\sim$4M \\

Nominal routes
& $\sim$1K \\

Route length, mean / median
& $49.79$ / $44.10$ m \\

Route length, P10--P90
& $31.47$--$76.01$ m \\

Route length, min--max
& $28.11$--$274.85$ m \\

Episode duration, mean / median
& $99.97$ / $91.3$ s \\

Episode duration, P10--P90
& $53.8$--$159.91$ s \\

Episode duration, min--max
& $35.6$--$396.5$ s \\

Simulation / sensing rate
& $10$ Hz \\

Route geometry families
& 6 \\

Follower motion programs
& 8 \\

Target-presence conditions
& 5 \\

Language-conditioned formations
& 6 \\

Human appearance configurations
& 756 \\

Nominal interaction combinations
& $8\times5\times6=240$ \\

Referent settings
& single + multi-person \\

Referent interventions
& counterfactual + reassignment \\

Scene variation
& static + pedestrian interference \\

\bottomrule
\end{tabular}
\end{table}

Route length refers to the nominal-route arc length and excludes additional travel caused by local scene-dependent deviations.
Long episodes allow multiple interaction events to occur within a single rollout, exposing the policy to repeated adaptation and recovery while route execution continues.

The motion, presence, and formation dimensions define
$8\times5\times6=240$ nominal interaction combinations before route geometry, human appearance, referent configuration, and scene variation are considered.
These factors are composed subject to physical and semantic compatibility rather than exhaustively enumerated as a Cartesian product.

The reported statistics describe the current training instantiation rather than the capacity of LeadInfra.
Its route generator, interaction programs, formation specifications, actor assets, and scene recipes remain configurable, allowing additional training conditions to be generated without changing the infrastructure.

\section{LeadBench Evaluation Protocol}
\label{app:leadbench}

\subsection{Frozen Episode Manifest}
\label{app:bench_manifest}

LeadBench evaluates all methods under a frozen episode manifest.
Each episode is specified before policy execution so that different methods face the same task condition rather than policy-dependent environment variation.
The benchmark uses held-out environments, routes, human actors, and scene instantiations that are disjoint from LeadInfra training, while preserving the same task definition and interaction contract.

For each episode, the manifest fixes the nominal route, initial robot and human configuration, target and distractor actor assignments, natural-language instruction, requested formation, follower behavior and presence conditions, local scene-interference layout, and the event windows used for response evaluation.
It also fixes the runtime configuration and terminal conditions required for reproducible closed-loop evaluation.

The manifest defines the evaluation condition but does not expose privileged supervision to the policy.
Hidden follower behavior states, ground-truth target annotations, privileged detour paths, future interaction schedules, and expert actions are excluded from policy-visible inputs.
In multi-person evaluation, matched counterfactual rollouts preserve the underlying route, actors, interaction program, and scene configuration while changing the language-designated target, as detailed in Sec.~\ref{app:bench_counterfactual}.

\subsection{Counterfactual Multi-Person Evaluation}
\label{app:bench_counterfactual}

Multi-person cases use matched target-A/target-B counterfactual rollouts to evaluate whether the robot responds to the language-designated follower rather than to unrelated nearby people.
Within each matched pair, the nominal route, human appearances, initial layout, requested formation, behavior program, and scene configuration are kept fixed.
Only the language-designated target and the corresponding target/distractor roles used for evaluation are exchanged.
Thus, the same physical behavior should influence the robot when produced by the designated follower, but should be ignored when produced by a distractor.

Each matched target-A/target-B pair is first aggregated into a single logical multi-person case before benchmark-level statistics are computed.
Single-person rollouts directly form individual logical cases.
This prevents multi-person conditions from receiving additional weight simply because referent attribution requires two matched physical rollouts.

\subsection{Common Runtime and Fairness}
\label{app:bench_runtime}

LeadBench evaluates all methods under the same frozen episode conditions and shared execution environment.
For each episode, the environment initialization, robot embodiment, sensing setup, nominal robot route, follower reference route, follower appearance and behavior, scene configuration, and environment-side execution conditions are kept fixed across policies.
Methods may retain their native action representations, inference frequencies, and model-specific preprocessing.

Evaluation is fully closed loop: each policy output changes the subsequent robot state and therefore the observations available at the next decision step.
Fairness is enforced through matched task conditions, a common policy-visible information boundary, and the same execution environment, rather than by forcing identical internal model interfaces.
No evaluated method receives privileged simulator information such as ground-truth target state, hidden behavior labels, future interaction schedules, expert actions, or privileged detour paths.

\subsection{Metric Definitions and Aggregation}
\label{app:bench_metrics}

Successful robot leading requires joint task completion, preservation of the intended route, appropriate responses to the designated follower, and sustained satisfaction of the requested interaction geometry.
LeadBench therefore reports four complementary metrics:
Lead Success Rate (LSR), Route Fidelity (RF), Target Response Score (TRS), and Social Contract Score (SCS).

\paragraph{Lead Success Rate (LSR).}
LSR measures joint robot--follower task completion.
An episode is successful only if the robot and designated follower both reach their frozen terminal regions, their terminal separation lies within $[0.8,4.0]$ m, and the joint terminal condition is maintained for at least $0.5$ s without an unrecovered hard failure:
\[
\mathrm{LSR}_e
=
\mathbb{I}
\left[
R_e \land F_e \land G_e \land \neg H_e
\right].
\]
Here, $R_e$ and $F_e$ denote robot and follower arrival, respectively;
$G_e$ denotes satisfaction of the terminal separation and hold-duration requirements;
and $H_e$ denotes an unrecovered hard failure.
Both robot and follower terminal regions use a $1.0$ m arrival tolerance.
LSR therefore measures whether the robot completes the navigation task \emph{with} its designated follower rather than merely reaching the goal itself.

\paragraph{Route Fidelity (RF).}
RF measures preservation of the nominal route together with forward progress.
After compressing consecutive stationary samples, let $d_i$ denote the lateral deviation of retained robot position $i$ from the nominal route.
We define
\[
q_i^{\mathrm{route}}
=
\begin{cases}
1, & d_i\leq1.0~\mathrm{m},\\
\exp\!\left[-(d_i-1.0)/0.5\right], & d_i>1.0~\mathrm{m}.
\end{cases}
\]
Let $s_i$ denote the corresponding projected arc-length progress along the nominal route.
Forward consistency is
\[
q^{\mathrm{forward}}
=
\frac{1}{N-1}
\sum_{i=2}^{N}
\mathbb{I}
\left[
s_i-s_{i-1}\geq-0.05~\mathrm{m}
\right],
\]
and
\[
\mathrm{RF}
=
\left(
\frac{1}{N}
\sum_{i=1}^{N}
q_i^{\mathrm{route}}
\right)
q^{\mathrm{forward}}.
\]
Compressing consecutive stationary samples prevents appropriate follower-aware waiting from accumulating an artificial route penalty.
RF therefore evaluates route preservation rather than movement alone.

\paragraph{Target Response Score (TRS).}
TRS measures whether interaction events produce the appropriate \emph{realized closed-loop robot response}, independent of the policy's internal action representation.
Each frozen event specifies its source (target or distractor), expected response
$c\in\{\mathrm{lead},\mathrm{slow},\mathrm{wait},\mathrm{resume}\}$,
onset, and reaction grace period.
Following the grace period, the realized robot behavior is evaluated over a $2.0$ s response window.

For the requested distance interval
$\mathcal{D}_{c_d}=[D_{\min},D_{\max}]$, define the distance error
\[
e(d_t,\mathcal{D}_{c_d})
=
\max
\left(
D_{\min}-d_t,\;
0,\;
d_t-D_{\max}
\right).
\]
Let $e_{\mathrm{pre}}$ and $e_{\mathrm{post}}$ denote the mean distance errors over the final $0.5$ s before the event and the final $0.5$ s of the response window, respectively.
Let $\Delta p$ denote non-negative route progress during the response window and $T$ its valid duration.
We use a reference progress speed
$v_{\mathrm{ref}}=0.5~\mathrm{m/s}$
and waiting displacement scale
$d_{\mathrm{wait}}=0.5~\mathrm{m}$.
The response-quality terms are
\[
Q_d
=
\exp(-e_{\mathrm{post}}/0.75),
\]
\[
Q_c
=
\exp
\left[
-\frac{
\max(0,e_{\mathrm{post}}-e_{\mathrm{pre}})
}{0.75}
\right],
\]
\[
Q_p
=
\operatorname{clip}
\left(
\frac{\Delta p/T}{v_{\mathrm{ref}}},
0,
1
\right),
\qquad
Q_s
=
\exp(-\Delta p/d_{\mathrm{wait}}).
\]

The response-specific scores are
\[
\mathrm{TRS}_{\mathrm{lead}}
=
\mathrm{TRS}_{\mathrm{resume}}
=
\sqrt{Q_pQ_d},
\]
\[
\mathrm{TRS}_{\mathrm{slow}}
=
\sqrt{Q_dQ_c},
\qquad
\mathrm{TRS}_{\mathrm{wait}}
=
\sqrt[3]{Q_dQ_cQ_s}.
\]
Lead and resume therefore require both forward progress and appropriate target distance;
slow rewards maintaining rather than worsening the interaction error;
and wait additionally penalizes continued route motion.
Distractor-control events expect \emph{lead} and are scored with respect to the designated target, so unnecessary reactions to distractors reduce the realized response score.
Frames belonging to frozen scene-interference intervals are excluded from TRS computation.

Importantly, TRS depends on physical closed-loop consequences rather than model-specific Pace values, velocities, waypoints, or action tokens.
It can therefore compare policies with different internal action representations.

\paragraph{Social Contract Score (SCS).}
SCS measures sustained satisfaction of the requested distance and relative bearing during stable leading.
Let $\mathcal{V}$ denote the frozen set of formation-valid frames.
Then
\[
\mathrm{SCS}
=
\frac{1}{|\mathcal{V}|}
\sum_{t\in\mathcal{V}}
\mathbb{I}
\left[
v_t=1
\land
d_t\in\mathcal{D}_{c_d}
\land
\theta_t\in\mathcal{B}_{c_b}
\right],
\]
where $v_t$ denotes target visibility and
$\mathcal{D}_{c_d}$ and $\mathcal{B}_{c_b}$ follow the interaction contract defined in Sec.~\ref{app:infra_behavior}.

Formation-valid intervals are frozen before evaluation and exclude phases in which the scripted interaction intentionally disrupts formation.
Target visibility itself is not used to remove frames from $\mathcal{V}$; therefore, target loss caused by the evaluated policy remains an SCS failure rather than reducing the denominator.
TRS and SCS are complementary: TRS measures event-driven responsiveness, whereas SCS measures sustained interaction quality during stable leading.

\paragraph{Aggregation.}
For TRS, events are first averaged within each response class, after which the response classes present in the evaluation unit are macro-averaged.
This prevents frequent lead events from dominating rarer slow, wait, or resume events.
A scheduled event that cannot obtain a valid response window because the rollout terminates early receives zero score.

The main benchmark table reports the six
Core/Easy/Constrained $\times$ single-/multi-person conditions separately.
When a single aggregate value is required, as in the ablation studies, each metric is first computed within the six conditions and then equally averaged:
\[
M_{\mathrm{agg}}
=
\frac{1}{6}
\sum_{d\in\{\mathrm{Core},\mathrm{Easy},\mathrm{Constrained}\}}
\sum_{p\in\{\mathrm{S},\mathrm{M}\}}
M_{d,p}.
\]
For multi-person evaluation, matched target-A and target-B rollouts are first combined into one logical case as described in Sec.~\ref{app:bench_counterfactual}.

Table~\ref{tab:leadbench_metric_params} summarizes the fixed evaluator parameters used throughout LeadBench.

\begin{table}[t]
\centering
\small
\caption{
\textbf{Key evaluator parameters used throughout LeadBench.}
}
\label{tab:leadbench_metric_params}
\begin{tabular}{lc}
\toprule
\textbf{Parameter} & \textbf{Value} \\
\midrule
Robot/follower terminal tolerance & $1.0$ m \\
Terminal separation range & $0.8$--$4.0$ m \\
Terminal hold duration & $0.5$ s \\
RF full-score deviation & $1.0$ m \\
RF deviation decay scale & $0.5$ m \\
RF backward tolerance & $0.05$ m \\
TRS response window & $2.0$ s \\
TRS pre/post statistics window & $0.5$ s \\
TRS distance-error decay & $0.75$ m \\
TRS reference progress speed $v_{\mathrm{ref}}$ & $0.5$ m/s \\
TRS waiting displacement scale $d_{\mathrm{wait}}$ & $0.5$ m \\
\bottomrule
\end{tabular}
\end{table}

\subsection{Failure and Termination Conditions}
\label{app:bench_termination}

LeadBench applies the same frozen termination rules to all evaluated policies.
These rules identify successful completion and policy failures that terminate closed-loop execution, as summarized in Table~\ref{tab:leadbench_termination}.

\begin{table}[t]
\centering
\small
\setlength{\tabcolsep}{3pt}
\renewcommand{\arraystretch}{1.08}
\caption{
\textbf{Episode termination conditions used in LeadBench.}
}
\label{tab:leadbench_termination}
\begin{tabularx}{\linewidth}{@{}lX@{}}
\toprule
\textbf{Condition} & \textbf{Definition} \\
\midrule

Success
& joint terminal contract satisfied continuously for $0.5$ s \\

Sustained collision
& collision penetration depth $\geq0.10$ m for more than $20$ consecutive frames ($2.0$ s at $10$ Hz) \\

Terminal synchronization timeout
& follower fails to satisfy the terminal contract within $3.0$ s after robot arrival \\

Route stall
& less than $0.25$ m route progress over $20.0$ s outside valid waiting periods \\

Route timeout
& frozen episode-specific maximum duration exceeded \\

\bottomrule
\end{tabularx}
\end{table}

The route-stall timer is suspended during frozen interaction periods in which waiting or target absence is expected, so appropriate follower-aware waiting is not treated as a policy failure.
Voluntary policy-induced stopping receives no such exemption.
Episode time limits are fixed by the manifest and depend on route duration rather than using a single global timeout.

Additional behavioral conditions, including prolonged target loss, excessive rotation, reverse progress, large target separation, and overtaking, are retained for diagnostic and failure analysis but do not independently terminate an episode in the final protocol.
For LSR, any policy rollout terminated by the failure conditions above is treated as unsuccessful; infrastructure-level runtime failures are retried and excluded from policy evaluation.

\section{Experimental Details and Additional Results}
\label{app:experiments}

\subsection{Baseline Adaptation and Fairness}
\label{app:baseline_details}

We adapt representative baselines to LeadBench while preserving the core modeling paradigm of each reference method.
When official implementations or checkpoints are available, we build on them directly; otherwise, we implement task-adapted versions that retain the corresponding perception, interaction, and control structure while modifying only the interfaces required for robot-leading evaluation.
These baselines are intended to represent their respective modeling paradigms under a common robot-leading setting rather than to reproduce the original evaluation environments of the reference methods.

All trainable baselines use LeadInfra for task-specific training or fine-tuning when applicable, whereas the classical Fiore-style controller requires no learned leading policy.
For modular baselines, we use Grounding DINO Tiny~\cite{liu2023grounding} for language-conditioned person grounding and Depth Anything V2 Small~\cite{yang2024depth} where monocular depth cues are required.
Uni-NaVid~\cite{zhang2024uni} is initialized from its released checkpoint and fine-tuned on LeadInfra.
Table~\ref{tab:baseline_adaptation} summarizes the resulting implementations.

All methods are evaluated under the same frozen LeadBench conditions and policy-visible information boundary described in Appendix~\ref{app:bench_runtime}.
No baseline receives additional ground-truth target state, hidden behavior labels, expert actions, future interaction schedules, or privileged detour information at test time.
Method-specific perception modules, action representations, inference schedules, and preprocessing are otherwise retained when compatible with the common LeadBench runtime.

\begin{table}[t]
\centering
\footnotesize
\setlength{\tabcolsep}{3.0pt}
\renewcommand{\arraystretch}{1.10}

\caption{
\textbf{Baseline implementations and LeadBench adaptations.}
Each baseline preserves the core modeling paradigm of the corresponding method while introducing the interfaces required for robot-leading evaluation.
}
\label{tab:baseline_adaptation}

\begin{tabularx}{\linewidth}{
@{}
>{\raggedright\arraybackslash}p{1.75cm}
>{\raggedright\arraybackslash}p{2.15cm}
>{\raggedright\arraybackslash}X
>{\centering\arraybackslash}p{1.25cm}
@{}
}
\toprule
\textbf{Method} &
\textbf{Paradigm} &
\textbf{LeadBench adaptation} &
\textbf{Training} \\
\midrule

Fiore et al.~\cite{fiore2015adaptive}
&
Classical human-aware guiding
&
Language-conditioned person grounding and temporal tracking; explicit follower-state control with geometric route execution.
&
None
\\

\addlinespace[2pt]

Nanavati et al.~\cite{nanavati2019follow}
&
Learned human--robot interaction state
&
Perception-derived follower state is modeled temporally and used to modulate geometric route execution.
&
LeadInfra
\\

\addlinespace[2pt]

K\"astner et al.~\cite{kastner2022human}
&
Semantic-state navigation
&
Explicit follower and local-scene semantics condition a learned navigation policy over the nominal route.
&
LeadInfra
\\

\addlinespace[2pt]

Uni-NaVid~\cite{zhang2024uni}
&
Foundation video-language-action navigation
&
Initialized from the released checkpoint; front/rear observations are adapted to its video interface, route context is provided through language, and the model is fine-tuned on LeadInfra.
&
LeadInfra
\\

\addlinespace[2pt]

StarVLA~\cite{ye2026starvla}
&
Direct action-query VLA
&
Matched direct-trajectory architectural control using the same Qwen3-VL backbone, temporal observations, language/route inputs, 24-step horizon, LeadInfra training data, and optimization budget as LeadVLA, but directly predicting the complete waypoint trajectory without Path--Pace--Residual composition.
&
LeadInfra
\\

\bottomrule
\end{tabularx}
\end{table}

\subsection{Human Expert Protocol}
\label{app:human_expert}

\begin{figure}[t]
    \centering
    \includegraphics[width=\linewidth]{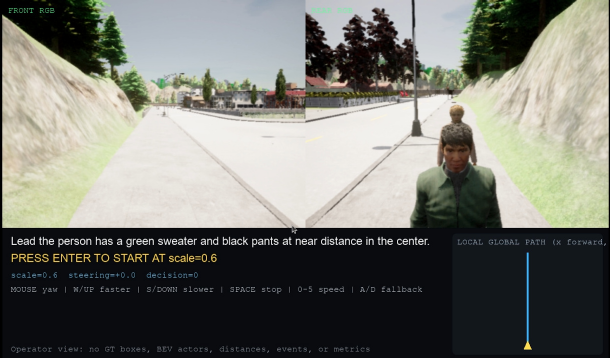}
    \caption{
    \textbf{Human-expert teleoperation interface.}
    Operators control the robot online while observing the same front/rear visual inputs, language instruction, and nominal route information available to evaluated policies.
    No simulator-privileged state is exposed through the interface.
    }
    \label{fig:human_teleop_ui}
\end{figure}

We additionally evaluate three human operators with robotics experience under the same frozen LeadBench episodes and policy-visible information boundary used for robot policies.
Each operator completes the full LeadBench evaluation through an online keyboard interface while observing only the front/rear visual observations, language instruction, and nominal route provided by LeadBench.
No ground-truth target state, hidden interaction label, future interaction schedule, or other simulator-privileged information is displayed.
The reported \textbf{Human Expert} results are obtained by averaging each LeadBench metric across the three operators.

Human Expert is reported as a reference rather than included in the robot-policy ranking.
Across the evaluated conditions, human performance remains substantially stronger on joint task completion and interaction-oriented metrics, while still falling below perfect TRS and SCS.
Human RF can also be slightly lower than that of strong route-following controllers because teleoperation introduces small deviations from the nominal route.
The human reference therefore provides a practical reference point under the same observation setting rather than a metric-perfect upper bound.
\subsection{Additional Ablation Analysis}
\label{app:additional_ablation}

\paragraph{Structured route execution across difficulty.}
The aggregate results in the main paper establish the overall benefit of structured route execution.
Table~\ref{tab:structured_difficulty} further reports LSR across LeadBench difficulty levels, averaging the single- and multi-person conditions within each level.

Replacing direct trajectory prediction with Path + Pace produces the largest gain under Core conditions, increasing LSR from $0.292$ to $0.646$.
This indicates that explicitly retaining the nominal route provides a strong execution prior when little local route recovery is required.
The role of Residual becomes more evident as route interference increases:
adding Residual changes Core LSR from $0.646$ to $0.604$, but improves Easy from $0.229$ to $0.312$ and Constrained from $0.188$ to $0.271$.

Together with the aggregate RF results in the main paper, this reveals the intended trade-off.
Path + Pace most strongly preserves nominal route execution, whereas Residual sacrifices some strict route fidelity in exchange for stronger task completion when local deviation and recovery become necessary.
The benefit of the full formulation therefore becomes more apparent under harder route-interference conditions rather than simply reflecting a uniformly more flexible trajectory decoder.

\begin{table}[t]
\centering
\small
\caption{
\textbf{LSR breakdown of action formulations across LeadBench difficulty levels.}
Single- and multi-person results are averaged within each difficulty level.
}
\label{tab:structured_difficulty}
\begin{tabular}{lcccc}
\toprule
\textbf{Action formulation} &
\textbf{Core} &
\textbf{Easy} &
\textbf{Constrained} &
\textbf{Overall} \\
\midrule

Direct Trajectory
& 0.292 & 0.167 & 0.063 & 0.174 \\

Path + Pace
& \textbf{0.646} & 0.229 & 0.188 & 0.354 \\

Path + Pace + Residual
& 0.604 & \textbf{0.312} & \textbf{0.271} & \textbf{0.396} \\

\bottomrule
\end{tabular}
\end{table}

\paragraph{Semantic learning and coupling across difficulty.}
\label{app:head_ablation}

The aggregate ablation in the main paper shows that semantic supervision and explicit semantic-to-action coupling serve different roles.
Table~\ref{tab:semantic_difficulty} further reveals how these effects vary with route-execution difficulty.

Auxiliary supervision alone performs strongly under Core conditions, increasing LSR from $0.646$ to $0.665$, but does not sustain this advantage as route interference increases, yielding $0.221$ on Easy and $0.101$ on Constrained.
This is consistent with the aggregate result in the main paper: semantic supervision can improve the shared follower-aware representation without necessarily improving overall task completion.

Shared-All Fusion exhibits a similar difficulty-dependent pattern.
It achieves the highest Core LSR of $0.667$ and improves Easy to $0.250$, but drops to $0.083$ under Constrained conditions.
Selective Fusion instead achieves the strongest Easy and Constrained results, with LSRs of $0.312$ and $0.271$, respectively, while maintaining competitive Core performance.

These results provide additional evidence for the control-role design underlying LeadVLA.
Semantic supervision is useful for shaping the shared representation, but directly coupling all predicted semantics to action does not remain robust as route execution becomes harder.
Selective coupling instead restricts the direct semantic pathway to information aligned with local trajectory adaptation, yielding the strongest overall performance.

\begin{table}[t]
\centering
\small
\caption{
\textbf{LSR breakdown of semantic-learning variants across LeadBench difficulty levels.}
Single- and multi-person results are averaged within each difficulty level.
}
\label{tab:semantic_difficulty}
\begin{tabular}{lcccc}
\toprule
\textbf{Variant} &
\textbf{Core} &
\textbf{Easy} &
\textbf{Constrained} &
\textbf{Overall} \\
\midrule

No Semantic
& 0.646 & 0.208 & 0.167 & 0.340 \\

+ Aux. Supervision
& 0.665 & 0.221 & 0.101 & 0.329 \\

+ Shared-All Fusion
& \textbf{0.667} & 0.250 & 0.083 & 0.333 \\

+ Selective Fusion
& 0.604 & \textbf{0.312} & \textbf{0.271} & \textbf{0.396} \\

\bottomrule
\end{tabular}
\end{table}

\paragraph{Referent-aware response attribution.}
\label{app:referent_analysis}

The aggregate results in the main paper show that referent-aware co-training improves closed-loop task performance.
We further examine whether robot behavior changes are attributed to the language-designated follower or to the distractor.

We identify behavior-change windows in which the robot changes its execution mode, such as slowing, waiting, or resuming, and associate each unambiguous window with the corresponding target or distractor interaction event.
We then report the fraction of attributable responses associated with each human role.
The two fractions are normalized over attributable response windows and therefore sum to $100\%$.
This attribution analysis is used only as a diagnostic and is separate from the TRS metric defined in Sec.~\ref{app:bench_metrics}.

As shown in Table~\ref{tab:referent_attribution}, without referent intervention, only $60.0\%$ of attributable behavior changes are associated with the designated follower, while $40.0\%$ are associated with distractor behavior.
Either intervention substantially improves target attribution:
Reassignment Only reaches $73.3\%$, while Counterfactual Only reaches $75.0\%$.
Combining both forms of supervision further increases target-attributed responses to $86.7\%$ and reduces distractor-attributed responses to $13.3\%$.

These results provide behavior-level evidence that referent-aware co-training shifts action modulation toward the language-designated follower.
The two interventions are therefore complementary not only in aggregate task performance, but also in reducing erroneous control attribution to surrounding people.

\begin{table}[t]
\centering
\small
\caption{
\textbf{Response attribution under different referent-aware training designs.}
Only behavior-change windows with an unambiguous target or distractor trigger are included.
Percentages are normalized over attributable response windows.
}
\label{tab:referent_attribution}
\begin{tabular}{lcc}
\toprule
\textbf{Training design} &
\shortstack{\textbf{Target-attributed}\\\textbf{responses $\uparrow$}} &
\shortstack{\textbf{Distractor-attributed}\\\textbf{responses $\downarrow$}} \\
\midrule

No Intervention
& 60.0\% & 40.0\% \\

Reassignment Only
& 73.3\% & 26.7\% \\

Counterfactual Only
& 75.0\% & 25.0\% \\

Full Co-training
& \textbf{86.7\%} & \textbf{13.3\%} \\

\bottomrule
\end{tabular}
\end{table}

\subsection{Analysis of Selective Semantic Coupling}
\label{app:semantic_coupling}

We further isolate the direct semantic pathways used by the two action branches.
These experiments distinguish two roles of semantic learning:
using auxiliary objectives to shape the shared action representation and explicitly injecting predicted semantics into an action branch.

\paragraph{Semantic coupling to Pace.}
Table~\ref{tab:pace_semantic_coupling} varies the semantic predictions explicitly provided to Pace while keeping the proposed lateral-and-formation coupling to Residual fixed.
The unfused Pace achieves the strongest overall LSR/SCS balance, and none of the additional semantic inputs provides a consistent improvement in task completion.
In particular, directly injecting distance, observation, or motion predictions does not improve LSR despite their apparent relevance to follower response.

These results suggest that Pace can exploit follower information through the shared multimodal action representation without requiring an additional semantic bottleneck.
Explicit semantic supervision remains useful for representation learning, but directly injecting its predicted labels into longitudinal route progress provides no consistent benefit.

\begin{table}[t]
\centering
\small
\caption{
\textbf{Effect of explicitly coupling semantic predictions to Pace.}
Residual uses lateral and formation semantics in all variants.
}
\label{tab:pace_semantic_coupling}
\begin{tabular}{lcccc}
\toprule
\textbf{Pace input} &
\textbf{LSR} &
\textbf{RF} &
\textbf{TRS} &
\textbf{SCS} \\
\midrule

None
& \textbf{0.396} & 0.853 & 0.352 & \textbf{0.250} \\

BBox
& \textbf{0.396} & 0.857 & 0.356 & 0.231 \\

Distance
& 0.333 & 0.861 & 0.353 & 0.235 \\

Observation
& 0.333 & 0.850 & \textbf{0.357} & 0.232 \\

Motion
& 0.333 & \textbf{0.868} & 0.349 & 0.243 \\

Obs. + Motion
& 0.333 & 0.857 & 0.346 & 0.229 \\

Dist. + Obs. + Motion
& 0.333 & 0.834 & 0.340 & 0.234 \\

All four
& 0.333 & 0.817 & \textbf{0.357} & 0.228 \\

\bottomrule
\end{tabular}
\end{table}

\paragraph{Semantic coupling to Residual.}
Table~\ref{tab:residual_semantic_coupling} keeps Pace unfused and varies only the semantic predictions explicitly provided to the Residual branch.
All variants retain the same Path--Pace--Residual formulation and auxiliary semantic supervision; the ``None'' variant removes only the direct semantic-fusion pathway.

With auxiliary supervision alone, the model reaches $0.329$ LSR and $0.180$ SCS.
Adding lateral semantics alone leaves LSR nearly unchanged at $0.331$, while improving RF to $0.900$ and SCS to $0.211$.
Formation coupling provides a clearer task-level improvement, reaching $0.354$ LSR and $0.238$ SCS.
Combining lateral and formation semantics yields the strongest task completion and interaction quality, with $0.396$ LSR and $0.250$ SCS.

The joint variant reduces strict RF relative to either single-cue variant while producing the largest gain in task completion, consistent with greater freedom for local spatial adaptation.
TRS remains broadly comparable across the variants and is not the primary beneficiary of this pathway.
These results support the proposed control-role alignment:
lateral relation and formation state provide complementary spatial cues when explicitly coupled to the Residual branch responsible for local trajectory adaptation.

\begin{table}[t]
\centering
\small
\caption{
\textbf{Effect of semantic coupling to Residual.}
Pace receives no explicit semantic fusion.
All variants retain the same auxiliary semantic supervision; ``None'' removes only direct semantic fusion to Residual.
}
\label{tab:residual_semantic_coupling}
\begin{tabular}{lcccc}
\toprule
\textbf{Residual input} &
\textbf{LSR} &
\textbf{RF} &
\textbf{TRS} &
\textbf{SCS} \\
\midrule

None
& 0.329 & 0.761 & \textbf{0.365} & 0.180 \\

Lateral
& 0.331 & \textbf{0.900} & 0.343 & 0.211 \\

Formation
& 0.354 & 0.885 & 0.346 & 0.238 \\

Lateral + Formation
& \textbf{0.396} & 0.853 & 0.352 & \textbf{0.250} \\

\bottomrule
\end{tabular}
\end{table}

Together, the Pace and Residual studies support the selective coupling used in LeadVLA.
All semantic objectives supervise the shared action representation, but explicit semantic-to-action fusion is most effective when the fused cue is aligned with the corresponding control role.
In our final design, Pace therefore remains unfused, while lateral and formation predictions are coupled explicitly to Residual.

\section{Real-World Evaluation and Qualitative Analysis}
\label{app:realworld}

\subsection{Real-Robot Setup and Evaluation Protocol}
\label{app:realworld_setup}

\begin{wrapfigure}{R}{0.38\textwidth}
    \centering
    \includegraphics[width=\linewidth]{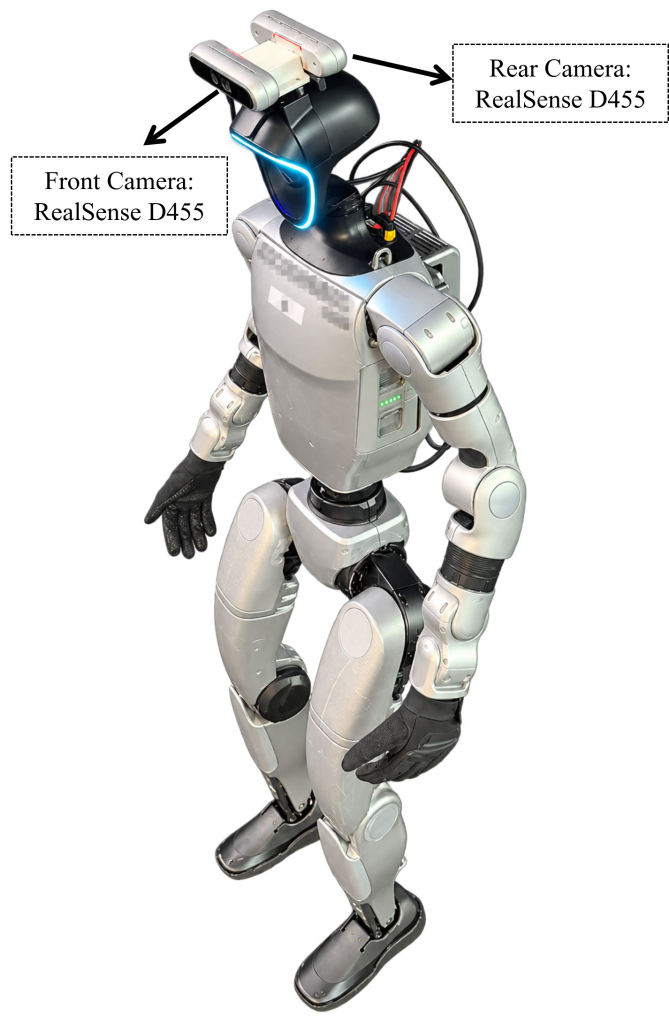}
    \vspace{-0.5em}
    \caption{
    \textbf{Real-robot platform.}
    LeadVLA is deployed on a Unitree G1 equipped with front- and rear-facing
    Intel RealSense D455 cameras.
    Only RGB observations are provided to the policy; depth is retained for
    offline evaluation.
    }
    \label{fig:realworld_hardware}
    \vspace{-0.9em}
\end{wrapfigure}

\paragraph{Robot and sensing.}
We deploy LeadVLA on a Unitree G1 humanoid equipped with front- and rear-facing
Intel RealSense D455 cameras, as shown in Fig.~\ref{fig:realworld_hardware}.
The policy interface is kept consistent with simulation: LeadVLA receives
front/rear RGB observations, the language instruction, and the frozen nominal
route.
Depth observations are not provided to the policy and are retained only for
offline evaluation of the Social Contract Score (SCS).
Physical locomotion is executed through the native Unitree G1 controller.

\paragraph{Global-route construction.}
For each environment, we first construct a map using SLAM.
A destination is then selected and a nominal global route is planned with ROS
\texttt{move\_base}.
Once generated, this route is frozen for the full episode and transformed into
the robot-local route representation required by LeadVLA at each policy query.
The nominal route is not replanned during policy execution.

\paragraph{Closed-loop deployment.}
Robot sensing and policy inference are connected through WebSocket communication.
At each policy query, the robot sends its current front/rear RGB observations
and route context to the inference server.
LeadVLA returns a 24-step robot-local waypoint trajectory, which is passed to
the robot-side trajectory-tracking interface and executed through the G1
locomotion controller.

The downstream controller tracks the trajectory predicted by LeadVLA but does
not generate an obstacle-conditioned replacement route or modify the frozen
nominal route.
In particular, no obstacle replanning is performed during policy execution.
Local deviations around route interference must therefore originate from the
trajectory predicted by LeadVLA rather than from a downstream navigation
planner.

Policy inference takes approximately $100$--$200$ ms on an NVIDIA RTX 5090
server.
WebSocket communication introduces approximately $100$ ms of additional
round-trip latency.
The resulting system operates fully closed loop: after each executed policy
trajectory segment, new visual observations and updated robot-local route
context are used for the next policy query.

\paragraph{Evaluation protocol.}
We evaluate four real-world settings corresponding to simplified
Core-S, Core-M, Easy-S, and Easy-M conditions from LeadBench.
Core evaluates follower-aware route execution without additional local route
interference, while Easy introduces local physical interference.
The multi-person settings additionally include a distractor.
Each setting contains 20 closed-loop trials, yielding 80 trials in total.
In each multi-person setting, the two participants each serve as the designated
follower in 10 trials.

Each episode contains two target-interaction events.
Their exact onset times are not fixed in advance, allowing the interaction to
occur naturally during execution while preserving the intended event type.
No real-world data are used for policy fine-tuning. 

We retain the same evaluation intent as LeadBench while adapting measurement to
the physical platform. The interaction-response component of SR is manually adjudicated from the recorded trials by checking whether the robot produces the expected response to the introduced target events. 
RF is computed from SLAM localization relative to the frozen nominal route.
TRS events and response windows are obtained through manual annotation of the
recorded trials.
SCS is computed offline using depth observations from the D455 cameras.

Because physical deployment uses a simplified success protocol rather than the
full LeadBench terminal contract, we report real-world \textbf{Success Rate
(SR)} separately from benchmark LSR.
A trial is counted as successful when the robot and designated follower both
reach the destination, the robot responds appropriately to the target
interaction events, and no obstacle collision occurs during the episode.

\subsection{Additional Real-World Analysis}
\label{app:realworld_analysis}

\begin{figure}[t]
    \centering
    \includegraphics[width=0.78\linewidth]{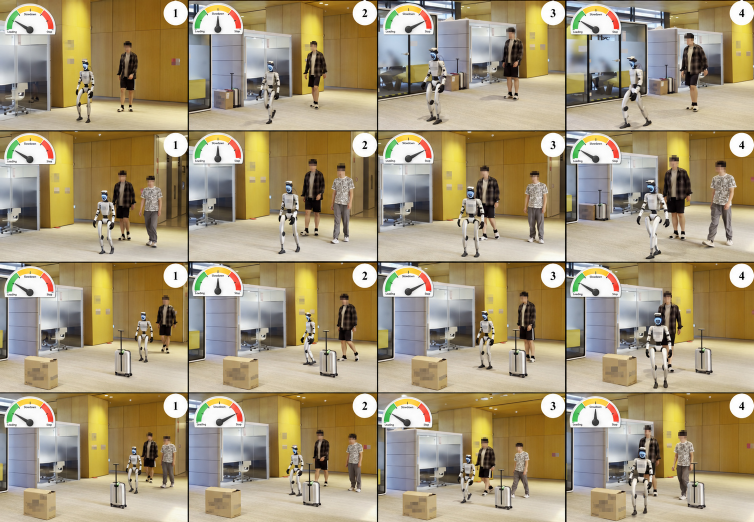}
    \caption{
    \textbf{Representative real-world closed-loop sequences.}
    Rows correspond to Core-S, Core-M, Easy-S, and Easy-M.
    The overlaid speed gauge visualizes changes in the robot's execution response over time.
    Across settings, LeadVLA largely preserves route-level progress while adapting its execution to the designated follower and local route conditions.
    }
    \label{fig:realworld_summary}
\end{figure}

\paragraph{Route fidelity remains strong after transfer.}
Figure~\ref{fig:realworld_summary} visualizes representative temporal sequences from all four real-world settings.
Across the physical trials, LeadVLA generally preserves the frozen nominal route while maintaining forward task progress despite visual and dynamic differences from simulation.
This observation is consistent with the relatively high RF values reported in the main paper ($0.80$--$0.95$), suggesting that the explicit Path reference provides a stable route-level execution anchor during sim-to-real transfer.

Performance nevertheless decreases from Core to Easy, where route execution additionally requires local spatial adaptation around interference.
Because RF remains comparatively high even in these settings, the remaining failures are associated less with wholesale loss of the nominal route than with the harder combination of local adaptation, follower response, and multi-person interaction.
Representative failure modes are discussed in Sec.~\ref{app:realworld_failure}.

\begin{figure}[t]
    \centering
    \includegraphics[width=0.78\linewidth]{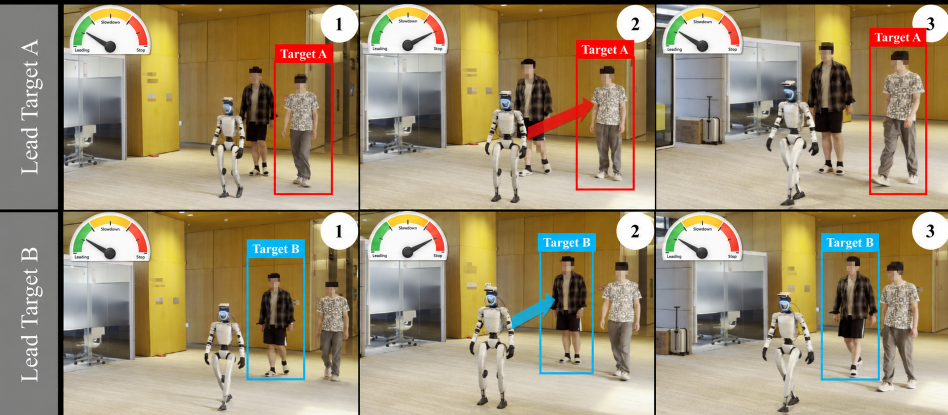}
    \caption{
    \textbf{Referent-conditioned control in the real world.}
    The physical setting and nominal route remain unchanged while the language instruction designates Target A (top) or Target B (bottom).
    LeadVLA changes its execution according to the currently designated follower, showing transfer of referent-aware control without real-world policy fine-tuning.
    Colored target annotations are added only for visualization and are never provided to the policy.
    }
    \label{fig:realworld_referent_switch}
\end{figure}

\paragraph{Referent-aware control is retained in physical deployment.}
We further examine whether follower-conditioned control remains associated with the language referent after sim-to-real transfer.
As shown in Fig.~\ref{fig:realworld_referent_switch}, we execute the same nominal route with both participants present while changing which participant is designated by the language instruction.
The robot correspondingly changes whose motion influences its execution, providing qualitative evidence that referent-aware behavior learned from LeadInfra transfers to physical deployment without real-world policy fine-tuning.

This behavior is most reliable when the two participants have visually distinct appearance cues.
In our qualitative observations, distinctive clothing makes target grounding easier, whereas visually similar target and distractor appearances can lead to ambiguous target binding and occasional responses to distractor motion.
Visual referent separability therefore remains an important limitation for multi-person real-world leading.

\subsection{Qualitative Observations and Failure Modes}
\label{app:realworld_failure}

\begin{figure}[t]
    \centering
    \includegraphics[width=0.78\linewidth]{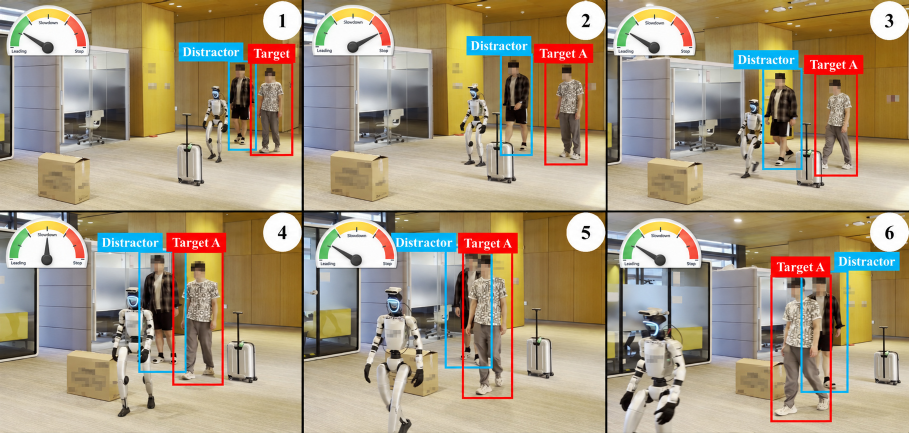}
    \caption{
    \textbf{Ignoring distractor behavior during real-world leading.}
    Target A remains the designated follower throughout the sequence.
    The distractor begins to lag, stops, and eventually falls substantially
    behind in the later frames, while LeadVLA continues leading Target A
    without slowing or waiting for the distractor.
    The speed gauge visualizes the robot's execution response over time.
    }
    \label{fig:realworld_distractor_ignore}
\end{figure}

\paragraph{Ignoring non-target behavior.}
Figure~\ref{fig:realworld_distractor_ignore} illustrates a long-duration
multi-person rollout in which the distractor progressively deviates from the
group interaction.
Although the distractor begins to lag and eventually stops, the robot continues
to execute the task with Target A.
This provides qualitative evidence that the learned policy can ignore salient
but task-irrelevant distractor behavior and maintain control attribution to the
language-designated follower in physical deployment.

\paragraph{Local spatial adaptation remains challenging.}
The largest degradation in the Easy settings occurs when local route
interference requires substantial deviation from the nominal route.
Although LeadInfra contains RGB-based supervision for local deviation and
recovery, real-world adaptation is often less decisive than in simulation,
and many observed Easy failures involve contact with the introduced obstacle.

Because the downstream controller does not perform obstacle replanning or
replace the frozen nominal route, successful local recovery must originate
primarily from the trajectory predicted by LeadVLA.
These failures therefore expose a remaining limitation in visually grounded
local trajectory adaptation under physical route interference.
Improving this capability while preserving follower-aware execution is an
important direction for future work.

\paragraph{Formation transfer is uneven.}
LeadVLA retains a visible notion of the requested interaction geometry after
sim-to-real transfer.
Changing near/far instructions produces observable changes in robot--follower
distance, which are also reflected in the depth-based SCS evaluation.
Lateral formation transfer is weaker:
when instructed to lead with the follower on the left or right, the physical
robot often keeps the follower closer to the center of the rear-view image
rather than maintaining a pronounced lateral offset.

This observation suggests that fine-grained lateral formation may require more
precise visual spatial reasoning than longitudinal distance regulation.
Because LeadVLA learns these relations directly from RGB observations without
real-world policy fine-tuning or additional geometric supervision, lateral
formation remains a challenging component of sim-to-real transfer.
We treat this as a qualitative observation rather than a separate quantitative
claim, as we do not perform a dedicated real-world lateral-formation study.

\paragraph{Physical deployment introduces additional domain gaps.}
Humanoid locomotion introduces camera vibration that is largely absent from the
simulation observations, adding temporal visual noise to both front and rear
streams.
Communication and inference latency can also delay the robot's response to
rapid changes in follower behavior.

Across the evaluated settings, route fidelity remains comparatively strong,
while finer-grained local adaptation, referent grounding, and lateral formation
become more challenging under physical deployment.
These observations suggest that future improvements should focus less on
recovering the nominal route itself and more on robustly adapting its execution
under real-world visual, interaction, and control uncertainty.

\section{Limitations and Future Work}
\label{app:limitations}

Despite relatively strong route fidelity, several aspects of follower-aware adaptation remain challenging.
First, local spatial adaptation is still limited when route interference must be handled jointly with follower response.
Our real-world experiments show that failures in the Easy settings are frequently associated with insufficient deviation and recovery around local interference, indicating that stronger visual spatial reasoning is needed to better coordinate route recovery with human-aware execution.
The current evaluation also focuses primarily on static or structured local interference; more general interaction with dynamic obstacles remains an important extension.

Second, fine-grained formation control transfers less reliably than route-level execution.
Distance-conditioned behavior remains observable in physical deployment, whereas left/right formation is more difficult to maintain.
This suggests that lateral formation places greater demands on precise visual spatial reasoning and geometry-aware control than coarse longitudinal distance regulation.

Third, referent-aware control remains sensitive to visual ambiguity.
The policy can condition its behavior on the designated follower when appearance cues are sufficiently distinguishable, but visually similar target and distractor appearances can still lead to ambiguous target binding and occasional distractor interference.
Harder appearance-matched counterfactuals, stronger temporal identity consistency, and richer multi-person interaction patterns may further improve referent attribution.

The current formulation also adapts primarily through locomotion.
A more capable robot leader could communicate proactively through viewpoint changes, body orientation, gestures, or verbal prompts, rather than relying only on slowing, waiting, and trajectory adjustment.
Extending the action space toward such communicative behaviors would allow future systems to study not only how the robot reacts to follower state, but also how it can actively shape the interaction.

More broadly, the remaining gap on LeadBench suggests that future robot-leading policies should move beyond purely reactive adaptation.
Predictive modeling of follower motion could allow the robot to anticipate lagging, formation drift, or temporary disengagement, and adjust its pace and local motion before the interaction becomes difficult to recover.
We therefore view stronger spatial reasoning, referent grounding, predictive human-motion modeling, and proactive communication as complementary extensions to the current structured formulation.
The central hypothesis remains that route-level task intent can be kept explicit and stable while learning progressively richer ways of adapting its execution to evolving human and scene conditions.

\stopcontents[appendix]

\end{document}